\documentclass[wcp]{jmlr}
\usepackage{multicol}
\usepackage{longtable}
\usepackage{booktabs}
\usepackage{setspace}

\usepackage{natbib}

\usepackage{mathtools} 
\DeclarePairedDelimiter{\ceil}{\lceil}{\rceil}
\usepackage{hyperref}
\hypersetup{
    colorlinks=true,
    linkcolor=blue,
    filecolor=magenta,      
    }
\usepackage{algorithm}
\usepackage{graphicx}
\usepackage{tikz}
\usepackage{pgfplots}
\newtheorem{assumption}{Assumption}

\DeclareMathOperator*{\argmin}{\arg\min}
\usepackage[utf8]{inputenc} 
\usepackage[T1]{fontenc}    
\usepackage{url}            
\usepackage{amsfonts}       
\usepackage{nicefrac}       
\usepackage{microtype}      
\usepackage{xcolor}         
\usepackage{lineno}

\usepackage{todonotes}

\usepackage{algcompatible,lipsum}
\usepackage{tikz}
\usepackage{pgfplots}

\usepackage{amsmath}
\usepackage{amssymb}
\usepackage{mathtools}
\usepackage{lineno}

\makeatletter
\let\Ginclude@graphics\@org@Ginclude@graphics 
\makeatother

\jmlrvolume{222}
\jmlryear{2023}
\jmlrworkshop{ACML 2023}

\title[Variance Reduced Pairwise Online Gradient Descent]{Variance Reduced  Online Gradient Descent for Kernelized \\ Pairwise Learning with Limited Memory}

 \author{\Name{Hilal AlQuabeh} \Email{hilal.alquabeh@mbzuai.ac.ae}\\
  \Name{Bhaskar Mukhoty} \Email{bhaskar.mukhoty@gmail.com}\\
  \Name{Bin Gu}* \Email{bin.gu@mbzuai.ac.ae}\\
 \addr Machine Learning Department, Mohamed bin Zayed University of Artificial Intelligence, Abu Dhabi, UAE}

\editors{Berrin Yan{\i}ko\u{g}lu and Wray Buntine}

\begin{document}

\maketitle

\begin{abstract}
Pairwise learning is essential in machine learning, especially for problems involving loss functions defined on pairs of training examples.
Online gradient descent (OGD) algorithms have been proposed to handle online pairwise learning, where data arrives sequentially. However, the pairwise nature of the problem makes scalability challenging, as the gradient computation for a new sample involves all past samples.
Recent advancements in OGD algorithms have aimed to reduce the complexity of calculating online gradients, achieving complexities less than $O(T)$ and even as low as $O(1)$. However, these approaches are primarily limited to linear models and have induced variance.
In this study, we propose a limited memory OGD algorithm that extends to kernel online pairwise learning while improving the sublinear regret. Specifically, we establish a clear connection between the variance of online gradients and the regret, and construct online gradients using the most recent  stratified samples with a limited buffer of size of $s$ representing all past data, which have a complexity of $O(sT)$ and employs $O(\sqrt{T}\log{T})$ random Fourier features for kernel approximation.  Importantly, our theoretical results demonstrate that the variance-reduced online gradients lead to an improved sublinear regret bound. The experiments on real-world datasets demonstrate the superiority of our algorithm over both kernelized and linear online pairwise learning algorithms. \footnote{The code is available at
\url{https://github.com/halquabeh/ACML-2023-FPOGD-Code.git}.} 
\end{abstract}

\section{Introduction}\label{sec:intro}
Pairwise learning is a machine learning paradigm that focuses on problems where the loss function is defined on pairs of training examples. It has gained significant attention due to its wide range of applications in various domains. For instance, in metric learning \cite{kulis2012metric}, pairwise learning is used to learn a similarity or distance metric between data points. In bipartite learning \cite{kallus2019fairness}, it is employed to address fairness concerns when making decisions based on two distinct groups. Multiple kernel learning utilizes pairwise learning to combine multiple kernels and enhance the performance of kernel-based methods \cite{gonen2011multiple}. AUC maximization involves pairwise learning to optimize the Area Under the ROC Curve, a popular evaluation metric for binary classification \cite{hanley1982meaning}. Pairwise differential Siamese networks utilize pairwise learning to compare and classify pairs of samples \cite{kang2018pairwise, song2019occlusion}.
\par 
Online pairwise learning is an effective approach for real-time decision-making, particularly when dealing with large-scale and dynamic datasets. The process involves sequentially processing data points and updating the model using pairwise examples. One technique that has been explored is online gradient descent, which provides computational efficiency and scalability. However, a drawback of online gradient descent in pairwise learning is its time complexity of $O(T^2)$, where $T$ represents the number of received examples. This is due to the requirement of pairing each new data point with all previous points, leading to significant computational complexity. To address this limitation, researchers have been investigating alternative methods such as buffering and sampling strategies \cite{zhao2011online,kar13,yang2021simple}. These approaches aim to reduce the computational burden and enable efficient learning on large-scale problems.
\par 
Online gradient descent has gained significant attention in the domain of online pairwise learning, leading to the development of various approaches. These methods, including online buffer learning \cite{zhao2011online,kar13}, second-order statistic online learning \cite{gao2013one}, and saddle point-problem methods \cite{ying2016stochastic,reddi2016stochastic}, have all employed linear models. Moreover, there has been limited exploration of non-linear models in this field, particularly with kernelized learning \cite{ying2015online,du2016online}.
One noteworthy approach in pairwise learning is online buffer learning, introduced by Zhao et al. \cite{zhao2011online}. This method utilizes a finite buffer with reservoir sampling to reduce the time complexity to $O(sT)$, where $s$ denotes the buffer size. By storing a subset of the data and ensuring uniform samples within the buffer, this technique effectively alleviates the computational burden. Furthermore, Yang et al. \cite{yang2021simple} achieved optimal generalization with a buffer size of $s=1$, marking a significant advancement in the field.
\par 
The existing frameworks in the literature have primarily focused on linearly separable data, overlooking the challenges associated with non-linear pairwise learning. Moreover, the online buffer methods proposed so far have not adequately addressed the sensitivity of generalization to the variance of the gradient. This limitation restricts their ability to capture the complexity present in real-world datasets.
Moreover, there is a lack of extensive research on non-linear pairwise learning, particularly in the context of kernel approximation. Although non-linear methods provide increased expressive power, the computational cost associated with kernel computation, which scales as $O(T^2)$ \cite{lin2017online,kakkar2017sparse}, poses challenges to their scalability and efficiency in practical applications.
In terms of generalization bounds, the analysis of online pairwise gradient descent with buffers and linear models has been extensively explored in previous works \cite{wang2012generalization, kar13}. These studies establish a bound of $O(1/s + 1/\sqrt{T})$ for this approach. However, it is important to note that this bound is only optimal when the buffer size $s$ is approximately $O(\sqrt{T})$, posing challenges for scenarios where a smaller buffer size is desired.
Additionally, the generalization analysis in \cite{yang2021simple} assumes independent examples in the sequential data, disregarding the temporal nature of the data and the potential ordering and correlation among data points. This assumption may lead to inaccurate performance estimation and unreliable convergence guarantees in online learning scenarios.
Taken together, these weaknesses highlight the need for further research and development in the field of online pairwise learning to address the limitations of linear frameworks, explore non-linear methods more comprehensively, and overcome the computational challenges associated with kernel computation.

\begin{table}[t]    
 \centering
 \caption{Recent pairwise learning algorithms(where $T$ is the iteration number, $d$: the dimension, $D$ is random features, V.R. denotes variance reduction, and $s$ is a buffer size), note the time complexity is w.r.t. gradients computations.}
\begin{tabular}{llllccc}\label{table:comparisons}\\
\hline
\textbf{Algorithm} 
& \textbf{Problem} & \textbf{Model} & \textbf{Scheme} & \textbf{V.R.} & \textbf{Time} & \textbf{Space} \\ \hline
\cite{gu2019scalable}             
& AUC          & Linear        & Offline           &  NA   &       $O(T)$      & $O(1)$         \\

\cite{pmlr-v80-natole18a}
& AUC              & Linear       & Online         & NA & $O(T)$         & $O(1)$                      \\

\cite{ying2016stochastic}
& AUC              & Linear       & Online         & NA   & $O(T)$         & $O(1)$                      \\

 \cite{zhao2011online} 
& AUC              & Linear         & Online            &  No  & $O(sT)$     & $O(s)$           \\
 \cite{gao2013one} 
& AUC              & Linear         & Online            &  No & $O(T)$     & $O(d)$           \\

    \cite{yang2021simple}             
& General          & Linear        & Online           &  No   &        $O(T)$     & $O(1)$          \\
\cite{kar13}             
& General          & Linear        & Online           &   No    &      $O(sT)$    & $O(s)$         \\
    \cite{lin2017online}             
& General          & Kernel        & Online           &   NA    &      $O(T^2)$      & $O(T)$       \\

 \cite{kakkar2017sparse}             
& AUC          & Kernel        & Offline           &   NA    &      $O(T\log{T})$    & $O(T^2)$         \\ 

\hline
FPOGD    (Ours)       
& General         & Kernel        & Online       & Yes    & $O(\frac{s D}{d} T)$        & $O(s)$                     \\ \hline
\end{tabular}
\end{table}

\par 
Our approach extends online pairwise learning to handle nonlinear data by incorporating kernelization of the input space. We address the impact of variance on regret through online stratified sampling, selectively updating the model based on cluster relevance. Utilizing random Fourier features, we efficiently estimate the kernel with sublinear error bound, achieving computational savings without sacrificing performance. By combining kernelization, efficient kernel approximation, and online stratified sampling, our method overcomes linear limitations, handles nonlinear data, and mitigates variance impact, resulting in a robust and effective online pairwise learning approach (Table \ref{table:comparisons}). Our main contributions can be summarized as follows:
\begin{itemize}
    \item We present an online pairwise algorithm for non-linear models with fast convergence. Our algorithm achieves sublinear regret with a buffer size of $O(s)$.
    \item We address variance impact on regret and propose online stratified sampling to control and improve the regret rate.
    \item For the case of Gaussian kernel, we approximate the pairwise kernel function using only $O(\sqrt{T}\log{T})$ features in comparison to $O(T)$ in previous works, while maintaining a sublinear error bound.
    \item We demonstrate the effectiveness of our proposed technique on numerous real-world datasets and compare it with state-of-the-art methods for AUC maximization. Our methodology showcases improvements across both linear and nonlinear models for the majority of the examined datasets.
\end{itemize}
The following sections are organized as follows. Section 2 introduces the problem setting, section 3 presents the proposed method, section 4 provides the regret analysis, section 5 discusses related work, followed by section 6 producing the experimental results, and finally section 7 concludes the paper.
 \section{Problem Setting}
The concept of pairwise learning arises in the context of a subset $\mathcal{X} \subset \mathbb{R}^d$ and a label space $\mathcal{Y} \subset \mathbb{R}$. It can be categorized into two cases:
\begin{itemize}
    \item  Pairwise hypothesis: This case involves learning a pairwise hypothesis, such as in metric learning, where the goal is to determine the relationship or distance between pairs of data points in $\mathcal{X}$. In particular, the hypothesis predicts the distance between pairs of instances i.e. $f:\mathcal{X}^2 \mapsto \mathcal{Y}$, and therefore given $n$ examples, the loss function is a finite sum of $\binom{n}{2}$ terms.

    \item Pairwise loss function: In this case, the focus is on minimizing a pairwise loss function, such as in AUC (area under curve) maximization. The objective is to optimize the ordering or ranking of pairs of data points based on their labels in $\mathcal{Y}$. In general the hypothesis is pointwise as in SVM, regression and binary deep classification,  i.e. $f:\mathcal{X} \mapsto \mathcal{Y}$, however the loss function itself represents the probability of predicting correctly the labels of opposing examples.
\end{itemize}
In our analysis, we specifically investigate pairwise loss functions from both branches. We establish a connection between the pairwise kernel associated with pairwise hypotheses and regular kernels. This enables us to explore the characteristics of the pairwise loss functions within the framework of regular kernels.
Consider an algorithm that learns from examples ${z_i := (x_i, y_i) \in \mathcal{Z} := \mathcal{X} \times \mathcal{Y}}$, where $i \in [T]$ denotes the number of examples. Let $f$ belong to space $\mathcal{H}$. In this paradigm, the pairwise loss function serves as a performance measure, denoted as $\ell: \mathcal{H} \times \mathcal{Z}^2 \rightarrow \mathbb{R}_+$.

Likewise, in online learning with pairwise losses, when a new data point $z_t$ is received, a local error is generated by incorporating the new data point together with all previous $t-1$ points.  The local error is then determined based on the chosen pairwise loss function as follows,
\begin{equation}\label{eq:localloss}
    {L}_t(f_{t-1}) =  \frac{1}{t-1} \sum_{i=1}^{t-1} \ell(f_{t-1},z_t,z_i) 
\end{equation}
The core objective in online pairwise learning is to create an ensemble of models, denoted as $f_1, f_2, \dots, f_T$, aimed at minimizing the expected risk. Assuming the data is mapped to a higher-dimensional space where linear separability is achieved, we consider a linear model represented as $w$. To address the issue of memory requirements, we employ a buffer-based local error denoted as $\bar{L}_t(w_{t-1})$, as defined in Equation \ref{eq:oneregret}. At each step $t$, the buffer, denoted as $B_t$, contains a limited number of historical example indices, and the cardinality of the buffer is represented as $|B_t|$ (equivalent to $s$ in the existing literature).
\begin{align}\label{eq:oneregret}
\bar{L}_t(w_{t-1})= \frac{1}{|B_t|}\sum_{i\in B_t}\ell(w_{t-1},z_t,z_i) 
\end{align}
The buffer plays a critical role in the learning process, being updated at each step using diverse strategies, ranging from randomized techniques like reservoir sampling \cite{zhao2011online,kar13} to non-randomized approaches like FIFO \cite{yang2021simple}. However, it is worth noting that there is a noticeable research gap regarding the variance implications of these sampling methods, despite their widespread utilization. 

To handle complex real-world data, our pairwise online approach assumes mapping both the hypothesis and the data to a Reproducing Kernel Hilbert Space (RKHS) denoted as $\mathcal{H}$. The associated Mercer pairwise kernel function $k:\mathcal{X}^4\mapsto \mathbb{R}$ satisfies the reproducing property $\langle k_{(x,x')}, g \rangle = g(x,x')$ , where $x, x' \in \mathcal{X}^2$ and $g \in \mathcal{H}$. In the case of pointwise hypothesis but pairwise loss functions, such as AUC loss, the kernel function simplifies to $k:\mathcal{X}^2\mapsto \mathbb{R}$. The space $\mathcal{H}$ encompasses all linear combinations of the functional mappings $\{k_{(x,x')}|(x,x')\in \mathcal{X}^2\}$ and their limit points.

To address the computational complexity of kernelization in the online setting, we utilize random Fourier features (RFF) as an efficient approximation of the Mercer kernel function. RFF provides a lower-dimensional mapping $r(\cdot)$, which approximates the kernel function, with the estimate denoted as $\bar{k}(\cdot)$. This approximation allows us to perform computations using linear operations, significantly reducing the computational complexity. The space spanned by the new kernel functions is denoted as $\hat{\mathcal{H}}$. Previous work has studied the error of random Fourier approximation in pointwise and offline settings. In the online setting, the minimum number of random features required to ensure sublinear regret has been found to be $O(T)$. In our method, we introduce an error bound for pairwise problems using only $O(\sqrt{T}\log{T})$ random features (see Section 5 for more details).

\subsection{Assumptions}
Before introducing our main theorems, we outline a set of widely accepted assumptions concerning the properties of the loss function and kernels. These assumptions hold significance in the realm of convex optimization and encompass commonly used loss functions such as squared loss as well as popular kernels like the Gaussian kernel.

\begin{assumption}[M-smoothness ]\label{ass:smmoth}
Assume for any $a\in\mathcal{Z}^2\times \mathcal{H}$, the gradient of the loss function $\nabla \ell(a)$ is M-Lipschitz continuous, i.e. $\forall w,w' \in \mathcal{H}$,
\begin{align}
    \nonumber \|\nabla \ell(a)-\nabla \ell(a')\|\leq M\left\|a-a'\right\|_2.
\end{align}
\end{assumption}

\begin{assumption}[Convexity]\label{ass:Convexity} 
Assume for any $z,z'\in\mathcal{Z}$, the loss function $\ell(\cdot,z,z')$ is convex function, i.e. $\forall w,w' \in \mathcal{H}$,
\begin{align}
    \nonumber \ell(w,z,z') \geq \ell(w',z,z') + \nabla \ell(w',z,z')^T (w - w').
\end{align}
\end{assumption}

\begin{assumption}[Finite Kernel]\label{ass:Kernel}
Assume for any $\rho$-probability measure on $\mathcal{X}^2$ the positive kernel function $k:\mathcal{X}^2 \times \mathcal{X}^2 \rightarrow \mathbb{R}$ is $\rho$-integrable, i.e. for any $(x,x') \in \mathcal{X}^2$,
\begin{align}
    \nonumber \int\int_{\mathcal{X}^2} k((x,x'),(\hat{x},\hat{x}'))d\rho(\hat{x})d\rho(\hat{x}')< \infty.
\end{align}
\end{assumption}

\subsection{Preliminaries}
In the analysis of buffer-based pairwise online gradient descent algorithms, two key concepts are essential for understanding the relationship between regret and variance in the proposed method.

\textbf{Variance of Stochastic Gradient.} Let us denote the variance of the stochastic gradient as $\mathbb{V}(u_t)$, with $u_t:=\nabla \bar{L}_t(w_{t-1})$ is the gradient based on a finite buffer. The variance is defined as the trace of the covariance matrix, i.e., $\mathbb{V}(u_t) := \mathbb{E}\|u_t - \mathbb{E}u_t\|^2$. The following lemma sheds light on the connection between the variance of the gradient and  the distance between inputs and their corresponding expected values.
\begin{lemma}\label{lemma:vartovar}
Assuming that the loss function is $M$-smooth, as mentioned in assumption \ref{ass:smmoth}, let $z_i$ denote the $i$-th sample drawn from a uniform distribution, i.e., $i \sim Uniform(1,t-1)$. Then, the variance $\mathbb{V}(u_t)$ of the stochastic gradient is bounded by:
        \begin{align}
         \mathbb{V}(u_t) \leq M^2  \mathbb{E}\|x_i-\mathbb{E}x_i\|^2 
    \end{align}
    where the expectation is w.r.t. the uniform random variable $i$.   Proof in appendix \ref{app:A1}.
\end{lemma}

\textbf{Regret Bound and Stochastic Gradient Variance.} The variance of the stochastic gradient plays a crucial role in determining the regret bound of pairwise online gradient descent algorithms. The following lemma establishes a connection between regret and the variance of the stochastic gradient. 

\begin{lemma}\label{theorem:regret 1}
With assumption \ref{ass:Convexity}, let $[w_t]_{i=1}^T$ be the sequence of models returned by running any buffer-based algorithm for $T$ time-steps using an online sequence of data. If $\bar{w}^* =\argmin_{w \in {\hat{\mathcal{H}}}} \sum_{t=2}^T {L}_t(w)$, and $B_t$ is sampled from the history of the received examples uniformly and independently, then the following hold,
\begin{align}
\sum_{t=2}^T {L}_t(w_{t-1}) &\leq   \sum_{t=2}^T {L}_t(\bar{w}^*) + 
  \frac{ \| \bar{w}^*\|^2}{2\eta} + \frac{\eta}{2} \sum_{t=2}^T(\mathbb{V}(u_t) + \|\mathbb{E}u_t\|^2)
\end{align}
where the expectation is w.r.t. the uniform distribution of buffer examples. The detailed derivation is in appendix \ref{app:A2}.
\end{lemma}
Hence, reducing the variance of the stochastic gradient can improve the regret of buffer online pairwise learning, which can be achieved using online stratified sampling as illustrated in next section.

\section{Proposed Method}
\begin{figure}[htbp]
    \centering
\includegraphics[width=\linewidth]{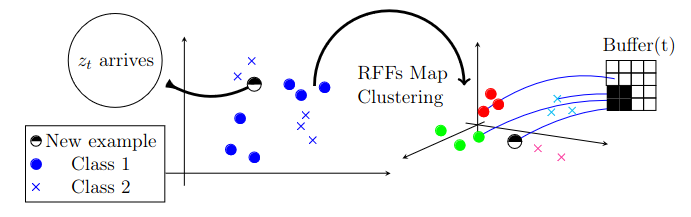}
    \caption{The incoming point $z_t$ from the left are transformed through kernelization and RFFs to a new space. Subsequently, they are clustered and added to the buffer using either a FIFO strategy or with a probability-based approach.}
    \label{figure:method}
    \vspace{-5mm}
\end{figure}

The proposed method consists of two essential parts that are mutually dependent. Firstly, by mapping non-linearly separable data to the RKHS $\mathcal{H}$, we achieve a transformation that renders the data linearly separable. This mapping serves as the foundation for effectively addressing non-linearity. Secondly, through the strategic implementation of stratified sampling, we potentially reduce the variance, preserve low memory utilization, and achieve sublinear regret.The initial mapping to the RKHS empowers us to seamlessly fulfill the objectives of the second component, ensuring an efficient approach overall as illustrated in figure \ref{figure:method}.

\subsection{Non-Linear Mapping to RKHS}
In pairwise online gradient descent algorithms, our goal is to handle non-linearly separable data. To achieve this, we employ a transformation $\Phi: \mathcal{X} \mapsto \mathcal{H}$ that maps the input space $\mathcal{X}$ to a high-dimensional RKHS $\mathcal{H}$. By performing this transformation, we project the non-linearly separable data into a higher-dimensional space where it becomes linearly separable.

However, the computational complexity associated with using explicit kernel computations can be prohibitive. To address this challenge, we leverage the RFFs technique, which allows us to approximate the inner products in the target space $\mathcal{H}$ more efficiently. Instead of directly computing the inner products in $\mathcal{H}$, we map the input space to an approximate space $\hat{\mathcal{H}}$ using a randomized mapping function. This approximation enables us to estimate the inner products of the original data points in the approximated space, rather than in the full high-dimensional space.

By applying this non-linear mapping using RFFs to the RKHS, we effectively handle the non-linearity in pairwise online gradient descent algorithms. This approach facilitates better separation of data points in the transformed space $\hat{\mathcal{H}}$, while maintaining a lower computational complexity of $O\left(\frac{sD}{d}T\right)$, compared to the $O(T^2)$ complexity associated with explicit kernel computations.

\vspace{-3mm}

\subsection{Online Stratified Sampling}

In addition to the non-linear mapping, we employ stratified sampling to further improve the efficiency and reduce the variance of the stochastic gradient estimates. Online Stratified Sampling (OSS) partitions the input space into balls of radius $\epsilon$, and ensures that each ball is represented by a uniform sample every iteration. By doing so, we achieve several advantages.

Firstly, stratified sampling reduces the variance of the stochastic gradient estimates. By partitioning the input space and uniformly sampling within each partition, we effectively minimize the variance of the stochastic gradient. This is achieved by reducing the expected distance between the sampled variables and their corresponding expected values, as highlighted in Lemma \ref{lemma:vartovar}.

Secondly, stratified sampling preserves low memory utilization. Instead of storing the entire history of received examples, we maintain one uniform sample from each partition at every step $t$. This approach reduces the memory requirement while still providing sufficient information to estimate the gradients accurately.

Finally, stratified sampling enables us to achieve sublinear regret. By reducing the variance and preserving low memory utilization, our method ensures efficient exploration and exploitation of the data, leading to improved regret bounds.
To this end, we redefine the loss at each time step by considering the presence of $\kappa_t$ partitions, denoted as $\mathcal{C}_j^t$, where each partition has a cardinality of $|\mathcal{C}_j^t|$, and the corresponding gradient is shown in equation \ref{newloss}.
\begin{align} \label{newloss}
  \nabla  \bar{L}_t(w_{t-1})=   \sum_{j =1}^{|B_t|} \frac{|\mathcal{C}_j^t|}{t-1} \nabla\ell(w_{t-1},z_t,z_{B_t[j]}) 
\end{align}
Note that the gradient mentioned above is unbiased, i.e. $\mathbb{E}\nabla  \bar{L}_t(w_{t-1}) = \nabla  {L}_t(w_{t-1})$. 
In order to reduce the variance in the stochastic gradient estimation, we maintain one uniform sample from each partition at every step $t$, ensuring that $|B|_t=\kappa_t$ . This update approach enables us to achieve lower variance. To accomplish this, we aim to find the optimal partitions at each time step $t$, which involves solving the following optimization problem or its upper bound based on Lemma \ref{lemma:vartovar}.

 \begin{align} \label{varUB}
      \min_{\mathcal{C}_j^t}  \quad  \mathbb{V} (u_t) &= \sum_{j=1}^{\kappa_t}\frac{(|\mathcal{C}_j^t|)^2}{ (t-1)^2} \mathbb{E}_{j|j\in\mathcal{C}_j^t} \| \nabla  \ell(w,z_t,z_j) - \mathbb{E}_{j|j\in\mathcal{C}_j^t} \nabla\ell(w,z_t,z_j)\|^2\\\nonumber
      &\leq \sum_{j=1}^{\kappa_t}\frac{(|\mathcal{C}_j^t|)^2M^2}{ (t-1)^2}  \mathbb{E}_{j|j\in\mathcal{C}_j^t}\|x_j-\mathbb{E}_{j|j\in\mathcal{C}_j^t}x_j\|^2 = \sum_{j=1}^{\kappa_t}\frac{|\mathcal{C}_j^t|M^2}{ (t-1)^2}  \sum_{j\in \mathcal{C}_j^t} \|x_j-\mathbb{E}_{j|j\in\mathcal{C}_j^t}x_j\|^2
    \end{align}
    
The objective in our approach bears resemblance to a conventional clustering problem, where $\mathbb{E}_{j|j\in\mathcal{C}_j^t}x_j=\sum_{j \in \mathcal{C}_j^t}\frac{x_j}{|\mathcal{C}_j^t|}$ represents the centroid of partition $j$ (note that the partition may be referred to as ``cluster'' intermittently). Our approach offers an effective solution by simultaneously addressing memory efficiency and variance reduction as certified in appendix \ref{app:A4}.

\subsection{The Algorithm}
 We introduce the algorithm in \ref{alg:FPOGD} for general pairwise learning with kernel approximation using RFFs.   
 The centroid update of the OSS algorithm involves minimizing the upper bound given in equation \ref{varUB}. This minimization can be achieved by utilizing the gradient of the upper bound with respect to the centroid $c$ of partition $j$, denoted as $c:=\mathbb{E}_{j|j\in\mathcal{C}_j^t}[x_j$]. The centroid update is then performed as follows: 
 \begin{equation}\label{updateCentroid}
     c\leftarrow c + \eta_c (z_t - c)
 \end{equation}
 where $z_t$ represents the newly assigned example to partition $j$, and $\eta_c$ is the step size. This update ensures that the centroid moves towards the assigned example in order to minimize the upper bound. In the subsequent section, we provide an analysis that decomposes the regret of the algorithm into two distinct components, first, the regret of learning $\bar{w}^* := \argmin_{w\in \hat{\mathcal{H}}} L(w)$, second, the regret of the kernel approximation using RFFs mapping. 


\begin{algorithm}[!t]
\caption{Fourier Pairwise Online Gradient Descent (FPOGD)}\label{alg:FPOGD}
  \begin{multicols}{2}	
	\begin{algorithmic}[1]
		\REQUIRE Initialization $w_1\in \hat{\mathcal{H}}$,  random feature size $D$, initial example $z_1$, Fourier feature distribution $p$ of kernel $k$, step size $\eta$, clustering distance $\epsilon$. 
        \STATE $B_1 \leftarrow \{z_1\}$, $C_1\leftarrow \{z_1\}$
		\FOR {$t = 2,\dots,T$}
		\STATE Receive new example $(x_t,y_t) \in \mathcal{Z}$
		\STATE Sample Fourier feature $\{q_i\}_{i=1}^{D/2}\sim p$
		\STATE Map to new space:
		 \[z_t =( \frac{2}{\sqrt{D}}[\sin(q_i^Tx_t),\cos(q_i^Tx_t)]_{i=1}^{D/2},y_t)\]
		\STATE Suffer loss \[g_t(w_{t-1}) = \sum_{j =1}^{|B_t|} \frac{|\mathcal{C}_j^t|}{t-1} \ell(w_{t-1},z_t,z_{B_t[j]})\]
		\STATE Select $v_t \in \partial g_t(w_{t-1})$
  	\STATE Update model $w_t = w_{t-1} - \eta v_t$
		\STATE Update $B_t,C_t = OSS(B_{t-1},C_{t-1},\epsilon,z_t)$

	\ENDFOR
 
    \STATE \textbf{Return} { $w_T$}
	\end{algorithmic}
 \columnbreak
	 \textbf{Online Stratified Sampling (OSS) with FIFO Buffer Update}
	\begin{algorithmic}[1]
		\REQUIRE Buffer $B$, Centroids $C$, Clustering threshold $\epsilon$, new example $z_t=(x_t,y_t)$
        \IF {$\min_{c\in C}\|x_t-c\|^2 \leq \epsilon$}
            \STATE Select cluster \[j:=\argmin_{c\in C}\|x_t-c\|^2\]
            \STATE Update buffer $B[j] \leftarrow  t$
            \STATE Update Centroid $j$ using equation \ref{updateCentroid}
        \ELSE 
        \STATE Add new cluster  $C \leftarrow C \cup t$
        \STATE Add new examples $B \leftarrow B \cup t$
        \ENDIF

     \STATE     \textbf{Return } $B,C$
	\end{algorithmic}


 
   \end{multicols}
\end{algorithm}

\section{Regret Analysis}
The regret of online algorithm relative to the optimal hypothesis in the space $\mathcal{H}$, i.e., $w^* = \argmin_{w\in \mathcal{H}} \sum_{t=2}^T L_t(w)$ when running on a sequence of $T$ examples is,
\begin{align}\label{eq:reg}
   \mathcal{R}_{w^*,T} = \sum_{t=2}^T L_t(w_{t-1}) - \sum_{t=2}^T L_t(w^*),
\end{align}
where the local all pairs loss $L_t(\cdot)$ is defined in equation \ref{eq:localloss}.
We can decompose the regret in equation \ref{eq:reg} by introducing  best-in-class hypothesis in the approximated space $\hat{\mathcal{H}}$, e.g:
$\bar{w}^*  = \argmin_{w\in \hat{\mathcal{H}}} \sum_{t=2}^T L_t(w)$ as follow:
\begin{align}\label{eq:regret}\nonumber
\mathcal{R}_{w^*,T} = \underbrace{\sum_{t=2}^T L_t(w_{t-1}) - \sum_{t=2}^T L_t(\bar{w}^*)}_{T_1}
+ \underbrace{\sum_{t=2}^T L_t(\bar{w}^*)  - \sum_{t=2}^T L_t(w^*)}_{T_2}
\end{align}
We provide the bound on $T_1$ in Theorem \ref{th:Allpairs} (Section \ref{sec4.2}), and then provide the bound on $T_2$ in Theorem \ref{regret_2} (Section \ref{sec4.3}). Finally, combining them together, we could provide the main theorem on the regret for the Algorithm \ref{alg:FPOGD} as follows.

\begin{theorem}
Let $\{z_t\in \mathcal{Z}\}_{t=1}^T$ be sequentially accessed by algorithm \ref{alg:FPOGD}. Let $D$ be the number of RFFs. And assumptions \ref{ass:smmoth}, \ref{ass:Convexity} and \ref{ass:Kernel} hold, with $M$ Lipschitz constant. Then, if the step size $\eta \in(0,1/M]$, the regret bound compared to $w^*$ is bounded with probability at least $1-2^8\left(\frac{\sigma R}{\delta}\right)\exp(-D\delta^2/(4d+8))$ as follow,
\begin{align}
    \mathcal{R}_{w^*,T} 
        \leq   \frac{M}{2} T  \|w^*\|^2_1 \delta^2   +\frac{\|\bar{w}^*\|^2 }{2\eta}+ \eta \sum_{t=2}^T\mathbb{V}({u}_t)
    \end{align}
    where $ \|w^*\|_1 = \sum_{i,j\neq i}^T|a^*_{i,j}|$, $\delta$ is the kernel approximation error, $\kappa_t$ is the number of clusters, $\sigma$ is the kernel width, $\mathbb{V}({u}_t) = tr(cov[u_t]) $, $R$ is the input diameter.
\end{theorem}
\begin{remark}
Choosing $\eta=\frac{1}{\sqrt{T}}$ and $\delta=T^{-1/4}$ makes the regret bound sublinear which is optimal. Moreover, if $\mathbb{V}(u_t)=0$ for all $t$'s, then it's possible to have $\log(T)$ regret by choosing $\eta=\frac{1}{\log(T)}$ and $\delta=\frac{\log(T)}{\sqrt{T}}$, which is similar to the case of full history update.Note that $D = O(\sqrt{T}\log(T))$ in general, but can be as low as $\log^2(T)$ for special kernels, (please refer to appendix \ref{app:B2}). 
\end{remark}

In the following, we provide the analysis to the upper bounds to $T_1$ and $T_2$ respectively.

\subsection{Regret in the Approximated Space $\hat{\mathcal{H}}$}\label{sec4.2}
The choice of buffer updating method, whether randomized (e.g., reservoir sampling) or non-randomized (e.g., FIFO), significantly impacts the analysis, as highlighted by \cite{kar13} and \cite{wang2012generalization}. To ensure independence between sampling randomness and data randomness, we begin with a simple FIFO approach, proving T1 bound in Theorem \ref{th:Allpairs} under the i.i.d. assumption. We then introduce reservoir sampling, which uniformly samples from the stream without $i.i.d.$ assumption, establishing convergence using the Rademacher complexity of pairwise classes. 

Consider the algorithm that has sequential access to the online stream . The following theorem demonstrates that the algorithm achieves optimal regret with memory complexity $O(\kappa_t)$.
\begin{theorem}\label{th:Allpairs}
With assumptions \ref{ass:smmoth} and \ref{ass:Convexity}, let $[w_t]_{i=1}^T$ be the sequence of models returned by running Algorithm \ref{alg:FPOGD} for $T$ times using the online $i.i.d.$ sequence of data. Then, if $\bar{w}^* = \arg\min_{w \in \hat{\mathcal{H}}} \sum_{t=2}^T L_t(w)$, and $\eta \in(0,1/M]$, the following holds:

\begin{align}
\sum_{t=2}^{T}L_t(w_{t-1}) - \sum_{t=2}^{T} L_t(\bar{w}^*) & \leq \frac{\|\bar{w}^*\|^2 }{2\eta}+ {\eta} \sum_{t=2}^T{\mathbb{V}({u}_t)} 
\end{align}
The proof is in appendix \ref{app:A3}.
\end{theorem}

\begin{remark}
    If the original space is assumed to be linearly separable (without kernels) then our algorithm has time complexity of $O(\kappa_T T)$ and offers sublinear regret with $\eta=1/\sqrt{T}$. Moreover note that for the case of $\epsilon \geq R$ the algorithm is equivalent to \cite{yang2021simple} and if $\epsilon\leq R/T$ it matches the algorithm in \cite{boissier2016fast}. In particular, if the clustering radius $\epsilon$ is large enough, it will result in only one cluster, similar to Yang's approach. Conversely, if $\epsilon$ is small (smaller than the distance between any two examples), it will create a cluster for each example, similar to Boissier's approach.
\end{remark}
\begin{remark}
    In the worst-case scenario, data points in pairwise learning are either assigned to new clusters or grouped within an epsilon distance from the initial centroid. The maximum number of clusters $\kappa_t$ at step $t$ can be upper-bounded by considering non-overlapping hyperspheres of radius epsilon in the bounded input space. Given the input space's volume $V_t^d$ at time step $t$, cluster threshold of $\epsilon$, and the gamma function $\Gamma(\cdot)$, we have:

 \begin{align}
      \kappa_t = \text{min} \, \bigg\{t-1,\ceil[\Big]{\frac{V_t^d \Gamma(d/2 + 1)} {\pi^{(d/2)}\epsilon^d}  }\bigg\}
 \end{align}

For a hypersphere input space with a constant radius $R$ at all time steps, the maximum number of clusters simplifies to $\kappa_t =min\, \{t-1,\ceil{(R/\epsilon)^d}\}$. In practice, the actual number of clusters obtained may be lower due to the data distribution (For experimental validations, please refer to Appendix \ref{app:B1}.).
\end{remark}
\textbf{Randomized Buffer Update}
In practical online learning scenarios, the assumption of an i.i.d. online stream is often impractical since the data can be dependent. Further, uniformly sampling from an online stream is not straightforward, making it challenging to achieve the bound mentioned earlier when the history examples are not readily available in a memory. To address this, we use buffer update strategies that force data independence in the buffer. Stream oblivious methods are particularly useful as they separate the randomness of the data from the buffer construction. To ensure effective buffer update and maintain the desired representation, we adopt reservoir sampling in conjunction with the clustering strategy. This approach treats each partition stream independently. When a new example arrives to cluster $\mathcal{C}_j^t$, the old example is replaced with a probability of $1/|\mathcal{C}_j^t|$, which makes it challenging to establish a uniform distribution among every cluster. Finally, the bound in theorem \ref{th:Allpairs} holds with the assumption of model-buffer independence (refer to appendix for further analysis).

\subsection{Regret of RFFs Approximation  }\label{sec4.3}
The kernel associated with the pairwise hypothesis in the space $\mathcal{H}$ is a function defined as  $k:\mathcal{X}^2\times \mathcal{X}^2 \mapsto \mathbb{R}^+$ with a shorthand $k_{(x,x')}(\cdot):=k((x,x'),(\cdot,\cdot))$  and can be constructed  given any uni-variate kernel $\mathcal{G}$ for any $x_1,x_2,x_1',x_2'\in\mathcal{X}$ as follow,
\begin{align}\label{eq:pairwiseKernel}
    k(x_1,x_2,x_1',x_2') &= \mathcal{G}(x_1,x_1') + \mathcal{G}(x_2,x_2')    - \mathcal{G}(x_1,x_2') - \mathcal{G}(x_2,x_1')  \\\nonumber
 &= \langle \mathcal{G}_{x_1} - \mathcal{G}_{x_2}, \mathcal{G}_{x_1'} - \mathcal{G}_{x_2'} \rangle_\mathcal{G}
\end{align}
It's clear that the pairwise kernel $k$ defined above is positive semi-definite on $\mathcal{X}^2$, and therefore it's Mercel kernel if $\mathcal{G}$ does on $\mathcal{X}$ (e.g. see \cite{ying2015online}). 
We further assume there exist a lower dimensional mapping $r:\mathcal{X}\mapsto R^D$, such that $ \mathcal{G}_{x}(\cdot) \approx r(x)^Tr(\cdot)$.

The quality of the approximation of the pointwise kernel $\mathcal{G}$ by random Fourier features is studied in literature (see \cite{rahimi2007random},\cite{bach2017equivalence},\cite{li2022sharp}), however the approximation of pairwise kernel $k(\cdot)$ needs further analysis. Let the kernel function $\mathcal{G}(\cdot)$ be shift invariant and positive definite, thus using Bochner's theorem, it can be represented by the inverse Fourier transform of a non-negative measure $p$ as $\mathcal{G}(x,x') = \mathcal{G}(x-x') = \int p(q) e^{i q^T(x-x')} du $ where $i=\sqrt{-1}$. For example if the kernel is the Gaussian kernel, the measure is found by Fourier transform to be $p \propto \mathcal{N}(\mathbf{0},diag(\sigma))$, where $\sigma \in \mathbb{R}^d$ is the kernel width. In other words, the kernel $\mathcal{G}$ can be approximated using Monte Carlo method, denoted as $\hat{\mathcal{G}}$, as follows:
\begin{align}
    \hat{\mathcal{G}}(x,x') = \frac{2}{D}\sum_{i=1}^{D/2} \cos(q_i^T(x-x')) = \langle r(x),r(x')\rangle
\end{align}
Where $q_i \sim \mathcal{N}(\mathbf{0},diag(\sigma))$, and  $r(x) := \frac{1}{\sqrt{D/2}}[cos(q_i^Tx),sin(q_i^T,x)]_{i=1}^{D/2}$. The following theorem bounds the random Fourier error in equation (\ref{eq:regret}).
\begin{theorem}
\label{regret_2}
Given a pairwise Mercer kernel $k_{(x,x')} :=k((x,x'),(\cdot,\cdot))$ defined on $\mathcal{X}^2\times\mathcal{X}^2$. Let $\ell(w,z,z')$ be convex loss that is Lipschitz smooth with constant ${M}$. Then for any $w^* = \sum_{i,j\neq i}^Ta_{i,j}^* k_{(x_i,x_j)}$, and random Fourier features number $D$ we have the following with probability at least $1-2^8\left(\frac{\sigma R}{\delta}\right)\exp(-D\delta^2/(4d+8))$,
\begin{align}
    \sum_{t=2}^T L_t(\bar{w}^*)  - \sum_{t=2}^T L_t(w^*) \leq \frac{M}{2} T  \|w^*\|^2_1 \delta^2
\end{align}
where $ \|w^*\|_1 = \sum_{i,j\neq i}^T|a^*_{i,j}|$. The proof is in appendix \ref{app:A5}.
\end{theorem}
 
 \begin{remark}
      Note that $\|w^*\|_1$ is controlled by the regularization, i.e. if there exist more than one optimal solution, then the optimal one has minimal $\|w^*\|_1$.
 \end{remark}

\section{Related Work}
Pairwise scalability poses a challenge in pairwise learning due to the quadratic growth of the problem with the number of samples. To address this issue, researchers have proposed different approaches in the literature. Some examples include offline doubly stochastic mini-batch learning \cite{dang2020large,gu2019scalable,alquabeh2022pairwise,alquabeh2022computational}, online buffer learning \cite{zhao2011online,kar13,yang2021simple}, second-order statistic online learning \cite{gao2013one}, kernelized learning \cite{hu2015kernelized,ying2015online}, and saddle point-problem methods \cite{ying2016stochastic,reddi2016stochastic}.
Online gradient descent, while having a time complexity of $O(T^2)$ \cite{boissier2016fast,gao2013one}, is impractical for large-scale problems. It pairs a data point $(x_t, y_t)$ received at time $t$ with all previous samples $\{x_{t'}, y_{t'} | 1 < t' < t-1\}$ to calculate the true loss. However, computing the gradients for all received training examples, which increases linearly with t, poses a significant challenge.
To address this, the work in \cite{zhao2011online} introduced two buffers, $B_+$ and $B_-$, of sizes $N_+$ and $N_-$, respectively, using Reservoir Sampling to maintain a uniform sample from the original dataset. While this approach provides a sublinear regret bound dependent on buffer sizes, it is limited to AUC maximization with linear models $(\mathcal{H} = \mathbb{R}^d)$ and overlooks the effect of buffer size on generalization error.
Researchers have also explored the application of saddle point-problem methods for tackling pairwise learning tasks involving metrics like AUC \cite{ying2016stochastic}. By formulating the problem as a saddle point problem and utilizing typical saddle point solvers, this approach achieves a time complexity of $O(T)$ in terms of gradient computations, providing an efficient solution for pairwise learning with reduced computational requirements.

\cite{kar13}, they introduce RSx, which replaces buffer samples with incoming data using $s$ Bernoulli processes at $1/t$, ensuring $s$ independent data points. However, achieving optimal generalization in terms of buffer loss requires a prohibitive buffer size of $O(\sqrt{T})$. Conversely, \cite{yang2021simple} argues for a buffer unity size, but their proof relies on impractical independent data streams and is limited to linear models. The common challenge in previous approaches is the need for $i.i.d.$ buffer examples for uniform convergence analysis and model-buffer decoupling. For instance, \cite{kar13} addresses this with modified reservoir sampling but still requires a buffer size of $O(\sqrt{T})$. Moreover, buffer-based loss becomes less informative as buffer updates become increasingly rare over time. These existing approaches suffer from limitations in computational efficiency, applicability to non-linear models, and a lack of explicit study of the impact of buffer variance on generalization error.

\section{Experiments}
We perform experiments on several real-world datasets, and compare our algorithm to both offline and online pairwise algorithms. Specifically, the proposed method is compared with different algorithms of AUC maximization, with the squared function as the surrogate loss.

\begin{table}[!t]
   	\small
 \center
\caption{AUC maximization results (average  $\pm$ standard error)$\times10^2$ using different batch and online algorithms on different datasets}
\begin{tabular}{lllllll}
{\textbf{Dataset}} & {\textbf{FPOGD}} & {\textbf{SPAM-NET}} & {\textbf{OGD}} & \textbf{S. Kernel} & \textbf{Proj++} & \textbf{Kar}\\ 
\hline 
diabetes  & {81.91}$\pm$0.48        & 82.03$\pm$0.32          & 82.53$\pm$0.31            & 82.64$\pm$0.37           & 77.92$\pm$1.44   & {79.85}$\pm$0.28      \\

ijcnn1                   & \textbf{92.32}$\pm$0.77                 & 87.01$\pm$0.10          & 83.46$\pm$1.25            & 71.13$\pm$0.59           & 92.20$\pm$0.27 & 83.44$\pm$1.21           \\

 a9a  & \textbf{90.03}$\pm$0.41  & 89.95$\pm$0.42 & 88.41$\pm$0.42           & 84.20$\pm$0.17 & 84.42$\pm$0.33  & 77.93$\pm$1.55 \\

mnist &\textbf{92.98}$\pm$0.38 &{88.57}$\pm$0.54 & 88.65$\pm$0.34& 89.21$\pm$0.15&89.82$\pm$0.15 &84.16$\pm$0.15\\

 rcv1   &\textbf{ 99.38}$\pm$0.20   & 98.13$\pm$0.15 &  99.05$\pm$0.57  & 96.26$\pm$0.35 & 94.54 $\pm$0.36 & 97.78 $\pm$0.64\\
  usps   & \textbf{95.02}$\pm$0.84   & 85.12$\pm$0.88 &  92.88$\pm$0.47  & 91.25$\pm$0.84 &90.14 $\pm$0.22 &  91.58$\pm$0.25 \\
    german   & \textbf{85.82}$\pm$0.24   & 76.89$\pm$2.46 &  84.20$\pm$0.54  & 80.11$\pm$0.44 &78.44 $\pm$0.66 &  84.21$\pm$0.45\\
\hline
Reg. & $l_2$  &    $l_1+l_2$  &    $l_2$ & $l_2$  &$l_2$& $l_2$ \\
\hline 
\end{tabular}
\label{table:AUC}

\end{table}
\subsection{Experimental Setup}

\begin{figure*}        [ht]
         \includegraphics[width=0.24\textwidth]{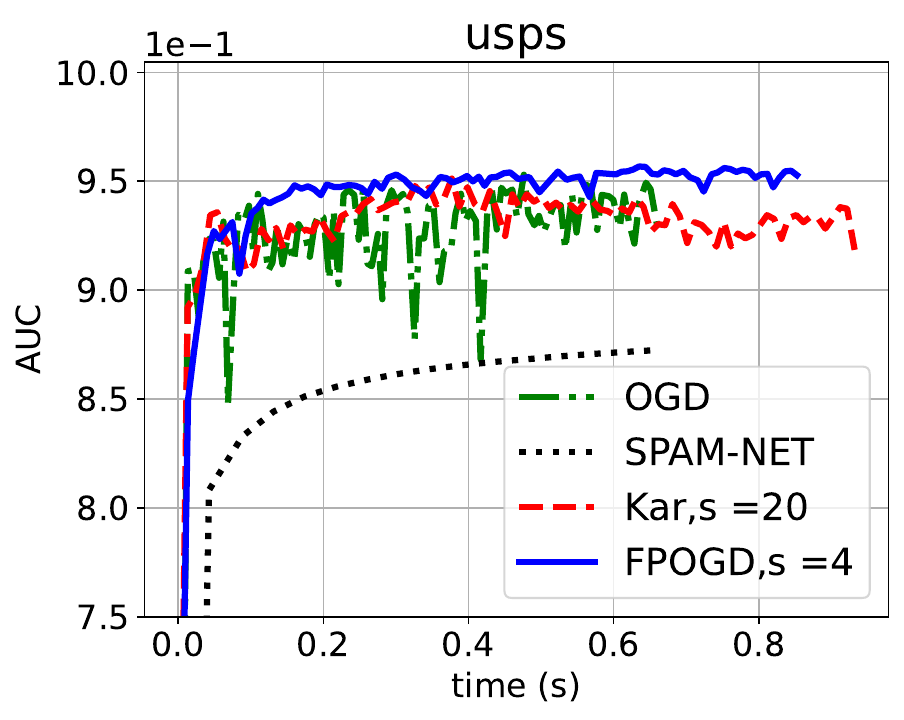}
        \includegraphics[width=0.24\textwidth]{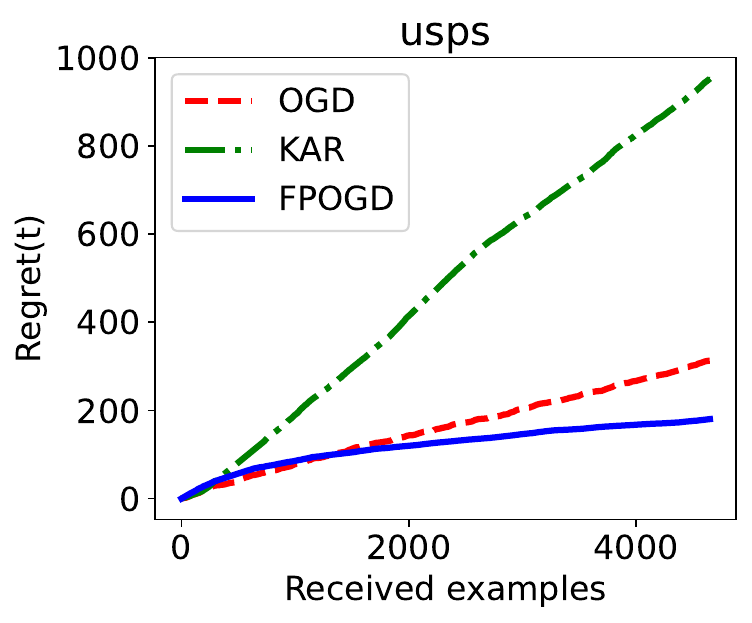}
        \includegraphics[width=0.24\textwidth]{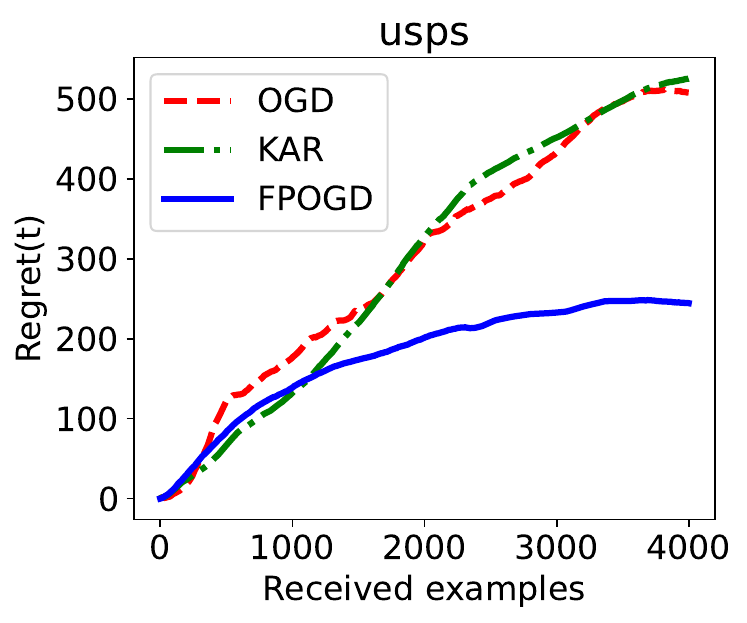}       
         \includegraphics[width=0.25\textwidth]{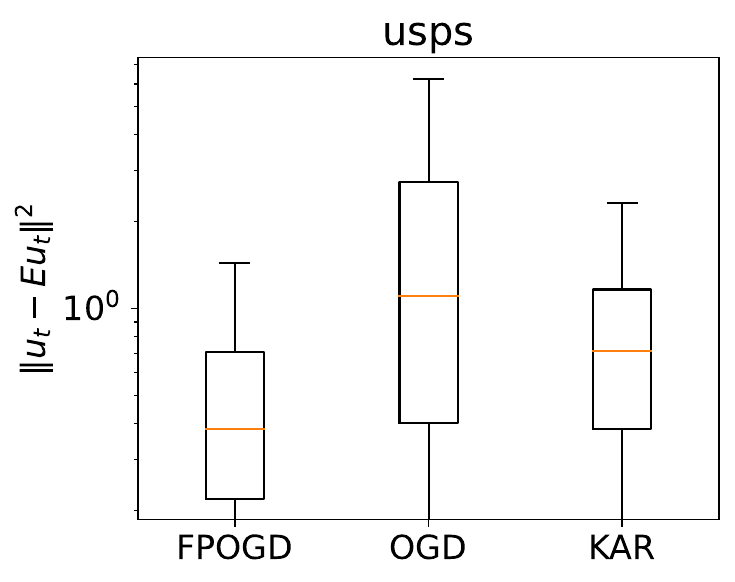}
         \includegraphics[width=0.24\textwidth]{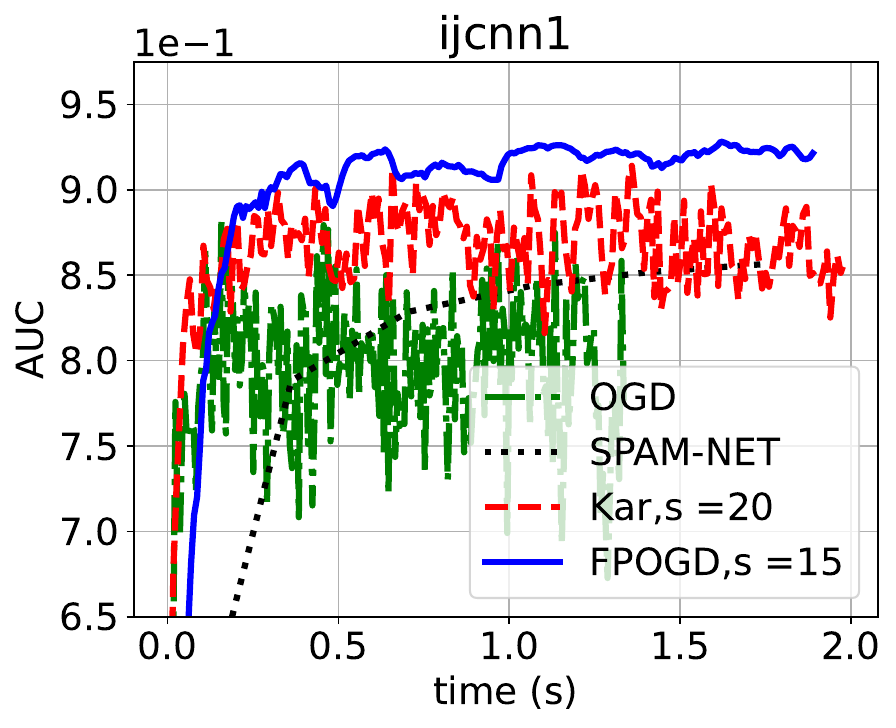}
        \includegraphics[width=0.24\textwidth]{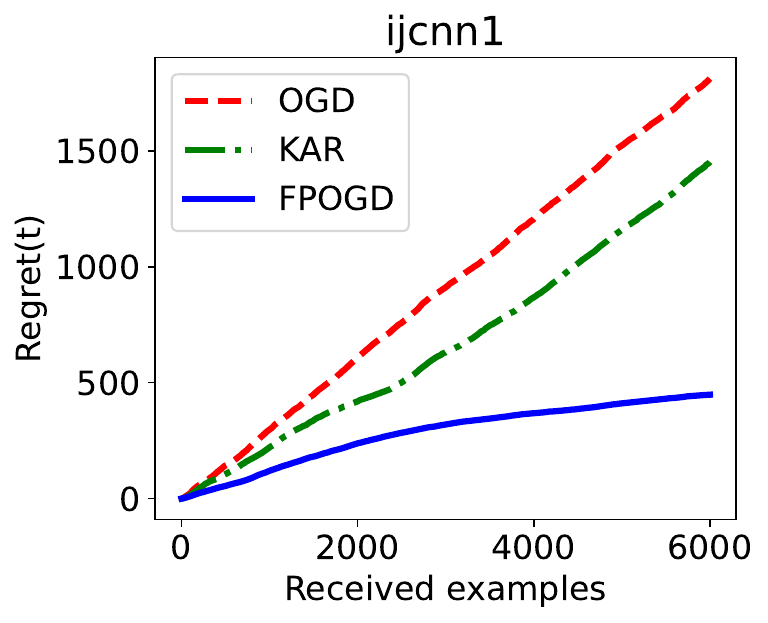}
        \includegraphics[width=0.24\textwidth]{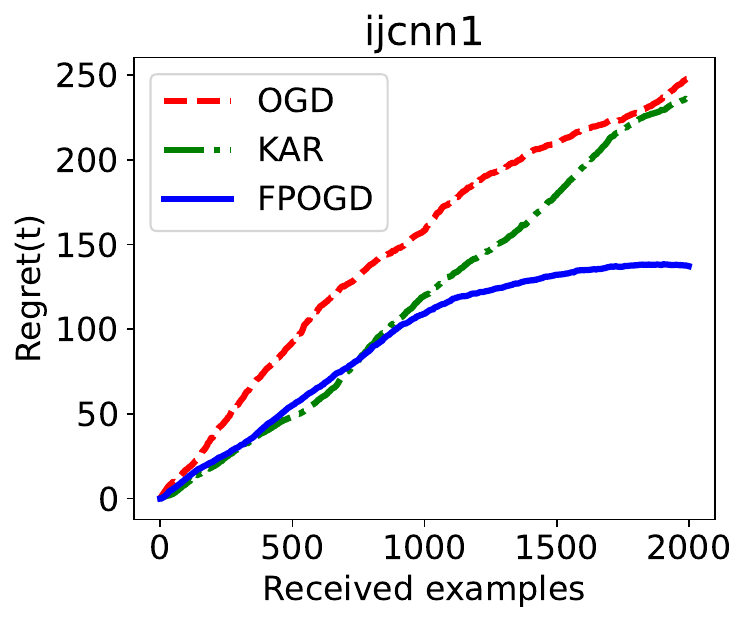}
         \includegraphics[width=0.25\textwidth]{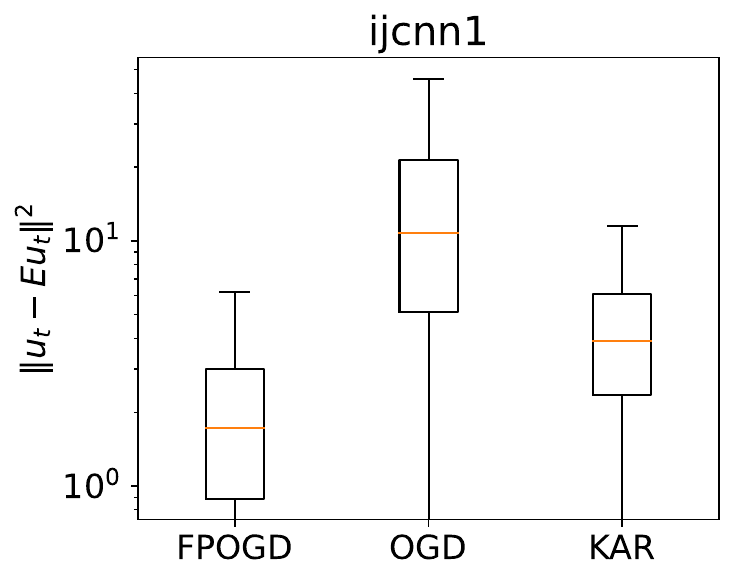}
     \caption{The first column presents AUC vs. time. Columns two and three display regret comparisons, with column two focused on i.i.d. datasets and column three on non-i.i.d. datasets. The fourth column provides insights into gradient variance analysis using a 4-size buffer, offering a glimpse into stochastic gradient behavior across buffer algorithms. Additional datasets are available in the appendix.}
     \centering
      \label{fig:results}
\end{figure*}
\paragraph{Compared Algorithms.} 
The compared algorithms includes offline and online setting are,
\begin{itemize}
    \item SPAM-NET \cite{reddi2016stochastic} is an online algorithm for AUC with square loss that is transformed into a saddle point problem with non-smooth regularization.
    \item OGD \cite{yang2021simple}, the most similar to our algorithm but with a linear model, that uses the last point every iteration.
    \item Sparse Kernel \cite{kakkar2017sparse} is an offline algorithm for AUC maximization that uses the kernel trick.
    \item Projection ++ \cite{hu2015kernelized} is an online algorithm with adaptive support vector set.
    \item Kar \cite{kar13} is an online algorithm with a randomized buffer update policy.
\end{itemize}
\paragraph{Datasets.} The datasets used in this study are sourced from the LIBSVM website \cite{CC01a}. Appendix B provides an overview of the dataset statistics, including the dataset name, size, feature dimension, and the ratio of negative to positive examples. Non-binary datasets undergo a conversion process into binary by evenly dividing the labels.

\paragraph{Implementation.} The experiments were validated for all algorithms through a grid search on the hyperparameters, employing three-fold cross-validation. For instance, in each algorithm, the step size, denoted as $\eta$, was varied within the range of $2^{[-8:-1]}$, providing flexibility for fine-tuning. Similarly, the regularization parameters, represented by $\lambda$, were explored over the range of $10^{[-8:-1]}$. In the case of the SPAM-NET algorithm, the elastic-net regularization parameter, denoted as $\lambda_2$, was determined through a grid search with values ranging from $10^{-8}$ to $10^{-1}$. To ensure a fair comparison, the use of kernelization is excluded when comparing with linear algorithms. All algorithms were executed five times on different folds using Python, running on a CPU with a speed of 4 GHz and 16 GB of memory.

\subsection{Experimental Results and Analysis}
The effectiveness of our random Fourier pairwise online gradient descent procedure in maximizing the area under the curve (AUC) is confirmed by our results obtained with a squared loss function as illustrated in figure \ref{fig:results}. Table \ref{table:AUC} clearly demonstrates that our algorithm outperforms both online and offline linear and nonlinear pairwise learning algorithms, yielding enhanced AUC performance particularly on large-scale datasets. Furthermore, in the appendix, we provide experimental results that demonstrate the relationship between the number of allowed clusters and the convergence of the algorithm. Additionally, we investigate the impact of the number of random features on the algorithm's performance.

\section{Conclusion}
In this research paper, we introduce a lightweight online kernelized pairwise learning algorithm. Our approach involves maintaining an online clustering mechanism and utilizing it to calculate the online gradient based on the current received sample. Additionally, we approximate the kernel function using random Fourier mapping. As a result, our algorithm achieves a gradient complexity of $O(sT)$ for linear models and $O(\frac{D}{d}sT)$ for nonlinear models, where $T$ represents the number of received examples and $D$ denotes the number of random features.


\begin{spacing}{0.15}
\bibliography{refs}
\end{spacing} 
\newpage
\appendix 
\section{Proofs}



\subsection{Proof of Lemma 1}
\label{app:A1}
\begin{proof}
Let $x_i$  sampled example from the history of examples, where $i\sim uniform[1,{t-1}]$,  then since the loss function is $M$-smooth we have,

\begin{equation}
    \nonumber \|\nabla \ell(w,z_t,(x,y))-\nabla \ell(w,z_t,(\mathbb{E}x,y))\|_2 \leq M \left\|x-\mathbb{E}x\right\|_2
\end{equation}
 then by adding and subtracting $\mathbb{E}\nabla \ell(w,z_t,z)$ to LHS after squaring both sides, and denote $\mathbb{E}z = (\mathbb{E}x,y)$ we have,

    \begin{align}\nonumber
        \Leftrightarrow&   \| \nabla \ell(w,z_t,z) - \mathbb{E}\nabla \ell(w,z_t,z) +\mathbb{E}\nabla \ell(w,z_t,z) -\nabla \ell(w,z_t, \mathbb{E}z)\|^2 \leq M^2 \|x-\mathbb{E}x\|^2\\\nonumber
        \Leftrightarrow&    \| \nabla \ell(w,z_t,z) - \mathbb{E}\nabla \ell(w,z_t,z) \|^2 + \|\mathbb{E}\nabla \ell(w,z_t,z) -\nabla \ell(w,z_t, \mathbb{E}z)\|^2 \\
       &   \quad  - 2 (\nabla \ell(w,z_t,z) - \mathbb{E}\nabla \ell(w,z_t,z))^T(\mathbb{E}\nabla \ell(w,z_t,z) -\nabla \ell(w,z_t, \mathbb{E}z)) \leq M^2 \|x-Ex\|^2
    \end{align}
    
taking expectation on both sides w.r.t. uniform distribution of $i$, and rearrange to have,
    \begin{align}\nonumber
        & \mathbb{E}  \| \nabla \ell(w,z_t,z) - \mathbb{E}\nabla \ell(w,z_t,z) \|^2 + \mathbb{E}\|\mathbb{E}\nabla \ell(w,z_t,z) -\nabla \ell(w,z_t, \mathbb{E}z)\|^2 \\\nonumber
        &   \quad  - 2 \underbrace{\mathbb{E}(\nabla \ell(w,z_t,z) - \mathbb{E}\nabla \ell(w,z_t,z))^T}_{=0}(\mathbb{E}\nabla \ell(w,z_t,z) -\nabla \ell(w,z_t, \mathbb{E}z)) \leq M^2 \mathbb{E}\|x-Ex\|^2\\\nonumber
        \Leftrightarrow   &\mathbb{E}\| \nabla \ell(w,z_t,z) - \mathbb{E}\nabla \ell(w,z_t,z) \|^2 \leq -  \|\mathbb{E}\nabla \ell(w,z_t,z) -\nabla \ell(w,z_t, \mathbb{E}z)\|^2 +M^2 \mathbb{E}\|x-Ex\|^2\\
           & \quad \quad  \quad \quad \leq M^2 \mathbb{E}\|x-Ex\|^2
    \end{align} 
Recall the definition of the variance of stochastic gradient to have final results. This completes the proof.
\end{proof}

\subsection{Proof of Lemma 2}\label{apd:first}
\label{app:A2}
\begin{proof}
Let $g_t(\cdot) =\frac{1}{|B_t|}\sum_{j\in B_t}\ell(\cdot,z_t,z_j) $ be convex function for all $t \geq 1$ where $B_t$ is the buffer of uniformly sampled $i.i.d.$ history examples. Let $u_t \in \partial g_t(w_{t-1})$. If we take the distance of two subsequent models to the optimal model we have, 
 	\begin{align}
 	\nonumber&   \| w_t - \bar{w}^* \|^2 - \| w_{t-1} - \bar{w}^* \|^2= \| w_{t-1} - \eta_t u_t -  \bar{w}^* \|^2 - \| w_{t-1} - \bar{w}^* \|^2\\\nonumber
 	&=\| w_{t-1} - \bar{w}^* \|^2-2 \eta_t u_t^T(w_{t-1}-\bar{w}^*) + \eta_t^2 \|u_t\|^2-\| w_{t-1}- \bar{w}^* \|^2\\ \nonumber
 	& = -2 \eta_t u_t^T(w_{t-1}-\bar{w}^*) + \eta_t^2 \|u_t\|^2\\
 	& \overset{}{\leq} -2 \eta ( g_t(w_{t-1}) - g_t(\bar{w}^*) )  +  \eta_t^2 \|u_t\|^2
 	\end{align}
 Where the last inequality implements Assumption \ref{ass:Convexity}, i.e. $ -u_t^T(w_{t-1}-\bar{w}^*) \leq -(g_t(w_{t-1}) - g_t(\bar{w}^*))$.

 Setting the step size $\eta_t=\eta$ for all $t$, and take the expectation w.r.t. the uniform randomness of the history points, and assume that if $w$ is fixed then $\mathbb{E}g_t(\cdot) = L_t(\cdot)$ 
	\begin{align}\label{results3}
	&L(w_{t-1}) - L(\bar{w}^*) 
\leq \frac{  \| w_{t-1}-\bar{w}^*\|^2 - \|w_t - \bar{w}^*\|^2}{2\eta}+\frac{\eta  \mathbb{E}\|u_t\|^2}{2} 
	\end{align}

Finally using the identity  $\mathbb{E}\|u_t\|^2 = \mathbb{V}(u_t) + \|\mathbb{E}u_t\|^2$, summing from $t=2$ to $t=T$ and setting $w_1=0$, would completes the proof.
\end{proof} 
\subsection{Proof of Theorem 5}
\label{app:A3}
\begin{proof}

 Starting from equation 4 with fact that the cluster-based buffer loss is unbiased of true local loss;
 	\begin{align}\label{results2}\nonumber
	&L(w_{t-1}) - L(\bar{w}^*) 
\leq \frac{  \| w_{t-1}-\bar{w}^*\|^2 - \|w_t - \bar{w}^*\|^2}{2\eta}+\frac{\eta  \mathbb{E}\|u_t\|^2}{2} 
	\end{align}
 and using the M-smoothness of the function $g$ i.e. $L(w_t) \leq L(w_{t-1}) + \mathbb{E}u_t^T(w_t-w_{t-1}) + \frac{M}{2} \|w_t-w_{t-1}\|^2$, and the update $w_t=w_{t-1} - \eta_t v_t$, 
 \begin{align}
     \mathbb{E}L(w_t) \leq \mathbb{E}L(w_{t-1}) - \eta \|\mathbb{E}u_t\|^2 + \frac{\eta^2M}{2} \mathbb{E}\|u_t\|^2
 \end{align}
By combining the last two inequalities and considering that the expectation is with respect to uniform sampling, we obtain the following inequality.
 \begin{align}\label{results}\nonumber
	& L(w_t) - L(\bar{w}^*)
\leq \frac{  \| w_{t-1}-\bar{w}^*\|^2 - \|w_t - \bar{w}^*\|^2}{2\eta} + (\frac{\eta }{2} + \frac{\eta^2M}{2})\mathbb{E}\|u_t\|^2 - \eta \|\mathbb{E}u_t\|^2 
	\end{align}
Using the fact that $\mathbb{V}(u_t) = \mathbb{E}\|u_t\|^2 -  \|\mathbb{E}u_t\|^2$ and choosing $\eta=(0,\frac{1}{M}]$, we have,
 \begin{align}\nonumber
	& L(w_t) - L(\bar{w}^*)
\leq \frac{  \| w_{t-1}-\bar{w}^*\|^2 - \|w_t - \bar{w}^*\|^2}{2\eta} + \eta \mathbb{V}(u_t)
	\end{align}
Finally summing from $t=2$ to $t=T$ and setting $w_1=0$ would complete the proof. 
\end{proof}
It is worth noting that our analysis remains valid in both scenarios: the FIFO buffer update, which requires independent examples, and the randomized update of the buffer which doesn't require online independent examples. While there is a coupling between the model $w_t$ and the buffer $B_t$, as the model $w_{t-1}$ incorporates information from the buffer at the previous step, we can still maintain the validity of the analysis by considering that this coupling is limited, as demonstrated in previous research (e.g., \cite{zhao2011online}). Although the gradient is not an unbiased statistic due to this coupling, we argue that the impact on the analysis is minimal.

Moreover, it is important to highlight that the main difference lies in the buffer size. In the case of coupling, the buffer size needs to be at least $\log{T}$, as determined through rigorous analysis utilizing techniques such as Rademacher complexity or covering number (for more details, refer to \cite{kar13} and \cite{wang2012generalization}). These analyses provide a deeper understanding of the underlying mechanisms and further support the validity of our approach.

\subsection{Certificate of Variance Reduction}
\label{app:A4}
    Assume that there exist $\kappa_t$ clusters, and denote $\hat{u}_t(\cdot)$ the cluster-based buffer gradient constructed using online stratified sampling 1, and ${u}_t(\cdot)$ represents the estimate obtained from uniform sampling without online clustering, 

\begin{align}\nonumber
   \mathbb{V} (u_t) &=  \frac{1}{ \kappa_t}\mathbb{E}_i \| \nabla  \ell(w,z_t,z_i) - \mathbb{E}_i \nabla\ell(w,z_t,z)\|^2 \\ \nonumber
    &=  \frac{1}{ \kappa_t}  \mathbb{E}_i \| \nabla  \ell(w,z_t,z_i)\|^2 -  \|\mathbb{E}_i \nabla\ell(w,z_t,z)\|^2  \\ \nonumber
    &\overset{(a)}{=} \frac{1}{ \kappa_t} \mathbb{E}_{\mathcal{C}_j^t} \mathbb{E}_{i|i\in \mathcal{C}_j^t} \| \nabla  \ell(w,z_t,z_i|i\in \mathcal{C}_j^t)\|^2 -  \|\mathbb{E}_i \nabla\ell(w,z_t,z)\|^2  \\ \nonumber
    & = \frac{1}{ \kappa_t} \sum_{j=1}^{\kappa_t} \frac{c_j^t}{t-1} \mathbb{E}_{i|i\in \mathcal{C}_j^t} \| \nabla  \ell(w,z_t,z_i|i\in \mathcal{C}_j^t)\|^2 -  \|\mathbb{E}_i \nabla\ell(w,z_t,z)\|^2  \\ \nonumber
    & =  \frac{1}{ \kappa_t}\sum_{j=1}^{\kappa_t} \frac{c_j^t}{t-1} \mathbb{V}_{i|i\in \mathcal{C}_j^t}  \nabla  \ell(w,z_t,z_i|i\in \mathcal{C}_j^t) + \|\mathbb{E}_{i|i\in \mathcal{C}_j^t} \nabla\ell(w,z_t,z_i|i\in \mathcal{C}_j^t)\|^2 -  \|\mathbb{E}_i \nabla\ell(w,z_t,z)\|^2  \\ \nonumber
& \overset{(b)}{=}  \frac{1}{ \kappa_t}\sum_{j=1}^{\kappa_t} \frac{c_j^t}{t-1} \mathbb{V}_{i|i\in \mathcal{C}_j^t}  \nabla  \ell(w,z_t,z_i|i\in \mathcal{C}_j^t) + \|\mathbb{E}_{i|i\in \mathcal{C}_j^t} \nabla\ell(w,z_t,z_i|i\in \mathcal{C}_j^t) - \mathbb{E}_i \nabla\ell(w,z_t,z)\|^2 \\
&{\geq}   \frac{1}{ \kappa_t} \sum_{j=1}^{\kappa_t} \frac{c_j^t}{t-1} \mathbb{V}_{i|i\in \mathcal{C}_j^t}  \nabla  \ell(w,z_t,z_i|i\in \mathcal{C}_j^t) = \mathbb{V} (\hat{u}_t)
\end{align}
    where equality (a) and equality (b) implements total expectation, i.e. $\mathbb{E}_{\mathcal{C}_j^t} \mathbb{E}_{i|i\in \mathcal{C}_j^t} \nabla\ell(w,z_t,z_i|i\in \mathcal{C}_j^t) = \mathbb{E}_i\nabla\ell(w,z_t,z_i)$.  The reduction in variance is influenced by the variances within each partition and the number of examples in it. If each cluster has same number of examples $(t-1)/\kappa_t$, we can observe that the bound becomes $1/\kappa_t$. Note that the maximum reduction in variance is ($t-1$), which is the case of full gradient. It is worth noting that the variance reduction assumes comparable variances among clusters. If the clusters have different variances, it is advisable to sample more from the high-variance cluster. This extension can be easily incorporated into our algorithm by considering the running variances of each cluster.

\subsection{Proof of Theorem 8}
\label{app:A5}
The study in \cite{bach2017equivalence} assumes that the true and approximated kernel functions belong to $L^2(\rho)$ i.e. space of square integrable functions (under the assumption \ref{ass:Kernel}), with the space $\mathcal{H}$ being dense in $L^2(\rho)$. Before proving the theorem, the following Corollary bounds the error of the pairwise kernel using main theorem in \cite{rahimi2007random}.
\begin{corollary}\label{RFerror}
Given $x_1,x_2,x_1',x_2'
\in\mathcal{X}$, and pairwise kernel $k$ defined on $\mathcal{X}^2\times\mathcal{X}^2$, the random Fourier estimation of the kernel has mistake bounded with probability at least $1-8\exp{\frac{-D^2\delta}{2}}$ as follow,
\begin{align}\nonumber
    \  |\hat{k}_{(x_1,x_2)}(x_1',x_2')   - {k}_{(x_1,x_2)}(x_1',x_2')  | \leq \delta  \end{align}
\end{corollary}
The proof follows from claim 1 in \cite{rahimi2007random} and the definitions of pairwise kernel $k$, which has four sources of errors.
\begin{proof}
\begin{align}\nonumber
 \sum_{t=2}^T L_t(\bar{w}^*)  - \sum_{t=2}^T L_t(w^*)&=  \sum_{t=2}^T\frac{1}{t-1}\sum_{i=1}^{t-1}\ell(\bar{w}^*,z_t,z_i) -\sum_{t=2}^T\frac{1}{t-1}\sum_{i=1}^{t-1}\ell({w}^*,z_t,z_i)  
 \\\nonumber
 &\overset{}{=}\sum_{t=2}^T\frac{1}{t-1}\sum_{i=1}^{t-1} \ell(\bar{w}^*,z_t,z_i) -\ell({w}^*,z_t,z_i) 
 \\
 &\overset{}{\leq} \sum_{t=2}^T\frac{1}{t-1}\sum_{i=1}^{t-1} \frac{M}{2} \|\bar{w}^* - {w}^* \|_2^2  
 \end{align}
 where last inequality applies assumption \ref{ass:smmoth}, and that fact that $\nabla \ell(w^*,z,z') = 0$ for any $z,z'\in \mathcal{Z}$. 
 Using the Representer theorem and the fact that the space $\mathcal{H}$ and $\hat{\mathcal{H}}$ are dense in $L^2(\rho)$ (space of squared integrable function, under assumption \ref{ass:Kernel}). Hence, we can approximate any function in $\mathcal{H}$ by a function in $L^2(\rho)$, i.e. without loss of generality we assume that $\bar{w}^* = \sum_{j=1,k\neq j}^{t-1} \alpha_{j,k}^* \hat{k}_{z_t,z_i}$, then we have $\|\bar{w}^* - {w}^* \|_2^2 
 \leq\|\sum_{j=1,k\neq j}^{t-1}\alpha_{j,k}^* (\hat{k}_{z_t,z_i} - {k}_{z_t,z_i} ) \|^2_{L^2(\rho)}$ using the fact that $\|\cdot\|_{L^2(\rho)}\geq \|\cdot\|_2$, finally using the triangle inequality we have,
 \begin{align}\nonumber
  \sum_{t=2}^T L_t(\bar{w}^*)  - \sum_{t=2}^T L_t(w^*) &{\leq}\sum_{t=2}^T \sup_{x_i,x_t\in\mathcal{X}} \frac{M}{2}\sum_{j=1,k\neq j}^{t-1} \|\alpha_{j,k}^* (\hat{k}_{z_t,z_i} - {k}_{z_t,z_i} )\|^2_{L^2(\rho)} \\ \nonumber
  &\overset{a}{\leq} \sum_{t=2}^T \frac{M}{2} \sum_{j=1,k\neq j}^{t-1} (\alpha_{j,k}^*)^2 \delta^2
  \\ \nonumber
 &\overset{b}{\leq} \sum_{t=2}^T \frac{M}{2}  \delta^2(\sum_{j=1,k\neq j}^{t-1} |(\alpha_{j,k}^*)|)^2 
 \\
 & = \frac{M}{2} T  \|w^*\|^2_1 \delta^2
 \end{align}
where inequality (a) implements corollary \ref{RFerror}, inequality (b) use the fact that sum of squares is less than the square of sum, and last equality assumes $ \|w^*\|_1 = \sum_{i,j\neq i}^T|a^*_{i,j}|$.
This completes the proof.
\end{proof}    
\section{Additional Experiments}
 \begin{table}[ht]
\centering
\caption{Datasets used in the experiments, where $N_-/N_+$ is the ratio of negative to positive examples.}
\begin{tabular}{llll}
Dataset     & Size    & Features & $N_-/N_+$ \\
\hline
diabetes      & 768   & 8 &34.90   \\
ijcnn1      & 141,691 & 22  & 9.45      \\
a9a         & 32,561  & 123  & 3.15    \\
MNIST    & 60,000 & 784  & 1.0     \\
covtype & 581012 & 54 & 1.0 \\
rcv1.binary & 20,242  & 47,236 &0.93   \\
usps & 9,298  & 256 & 1.0   \\
german & 20,242  & 24 & 2.3   \\
\hline
\end{tabular}
\label{table:datasets}
\end{table}
\begin{figure}        [ht]

  \begin{multicols}{2}
\label{Numbers}
  \includegraphics[width=0.98\linewidth]{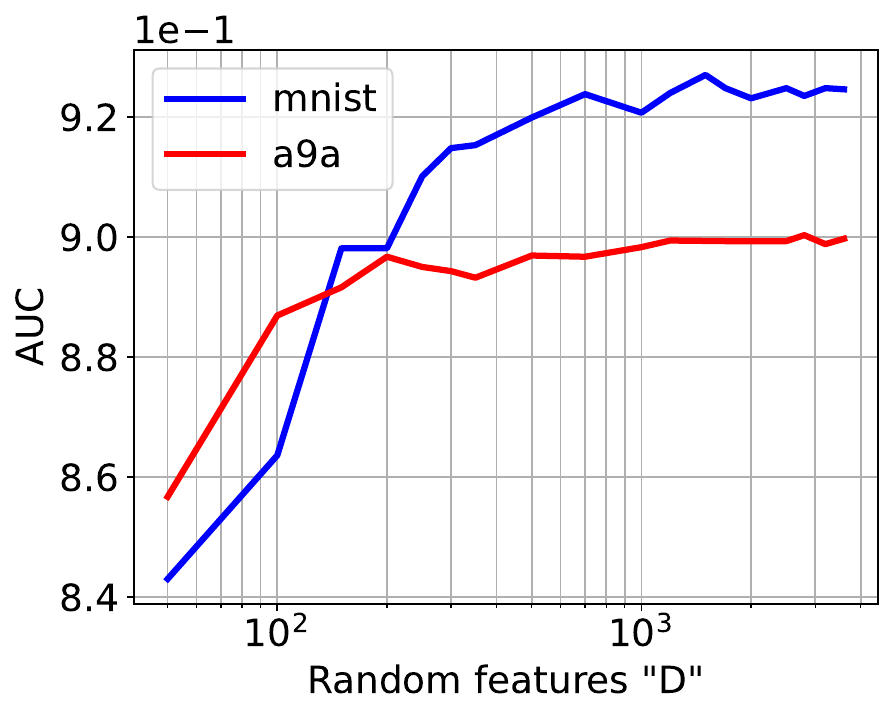}

 \columnbreak
  \includegraphics[width=0.98\linewidth]{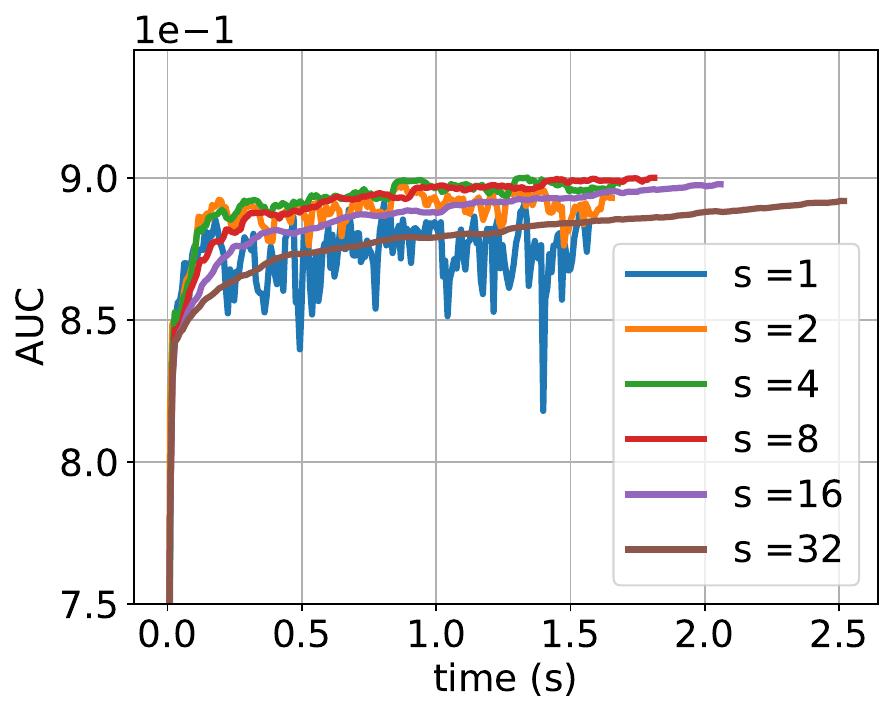}

 \end{multicols}
      \caption{ On left, AUC versus number of random features D used to approximate the kernel, for a9a and MNIST datasets. On right,AUC versus maximum number of clusters "s" in algorithm 1 for the dataset "a9a".}
\end{figure}
\subsection{Number of Clusters}\label{NC}
\label{app:B1}
In our algorithm, the number of clusters is determined by the hyperparameter epsilon. To manage memory requirements and computational costs, we enforce a maximum limit on the cluster number in our experiments, even if epsilon theoretically permits more clusters. This strategic limitation not only ensures resource efficiency but also contributes to practical applicability.

In the figure below, we shed light on the influence of varying cluster limits on the AUC score, specifically focusing on the "a9a" dataset while utilizing a small epsilon value. The findings reveal that as the number of clusters increases, the AUC score exhibits a gradual ascent, taking more time to reach its peak value. Importantly, these results also underscore that by constraining the number of clusters, we achieve a noticeable reduction in variance while still maintaining a commendable level of performance. This balance between performance and resource management is crucial for the practical implementation of our method.
\subsection{Number of Random Feature}\label{DD}
\label{app:B2}
By utilizing the Mercer decomposition theorem and the properties of eigenvalues, we can derive bounds on the number of random Fourier features needed for different decay rates of the eigenvalues. Specifically, for a decay rate of $1/i$, the sufficient number of features is $D\geq 5T\log{2T}$. For a decay rate of $R^2/i^{2c}$, the sufficient number of features is $D\geq T^{1/2c}\log{T}$. And for a geometric decay rate of $r^i$ ($r>1$), the sufficient number of features is $D\geq \log^2{T}$.

In our experiments, we focus on a Gaussian kernel with a constant width of $1/d$, where $d$ is the dimension of the input space. For this kernel, the required number of random features is $O(\sqrt{T}\log({T}))$. Figure 1 illustrates the results of our experiments, which show that $D=O(\sqrt{T}\log({T}))$ is sufficient for a good approximation of the kernel, as the AUC does not improve significantly beyond this point.
\subsection{Additional datasets}
\label{app:B3}
\begin{figure*}        [ht]

     \centering
         \includegraphics[width=0.24\textwidth]{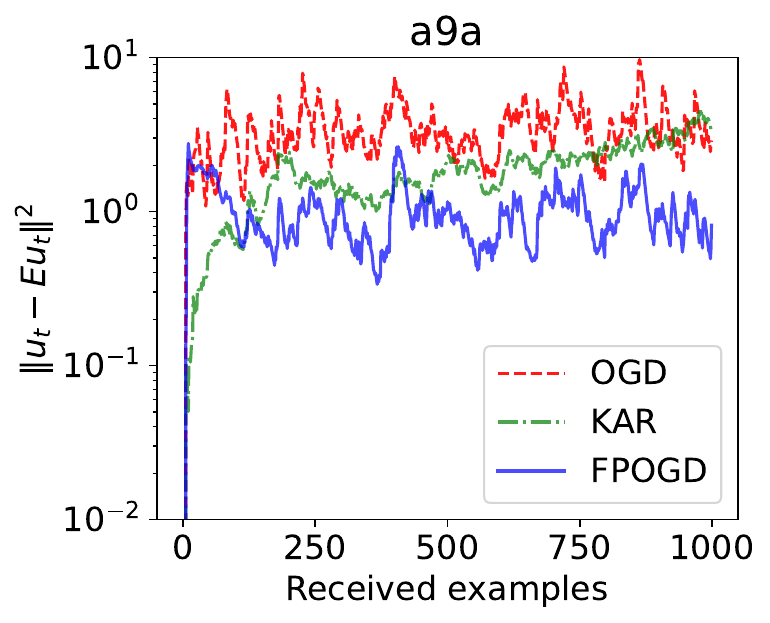}
         \includegraphics[width=0.24\textwidth]{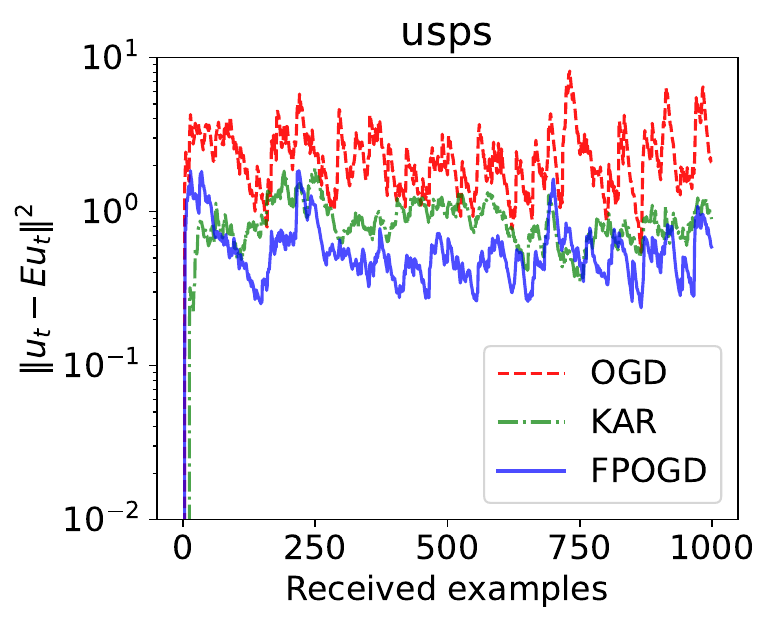}
         \includegraphics[width=0.24\textwidth]{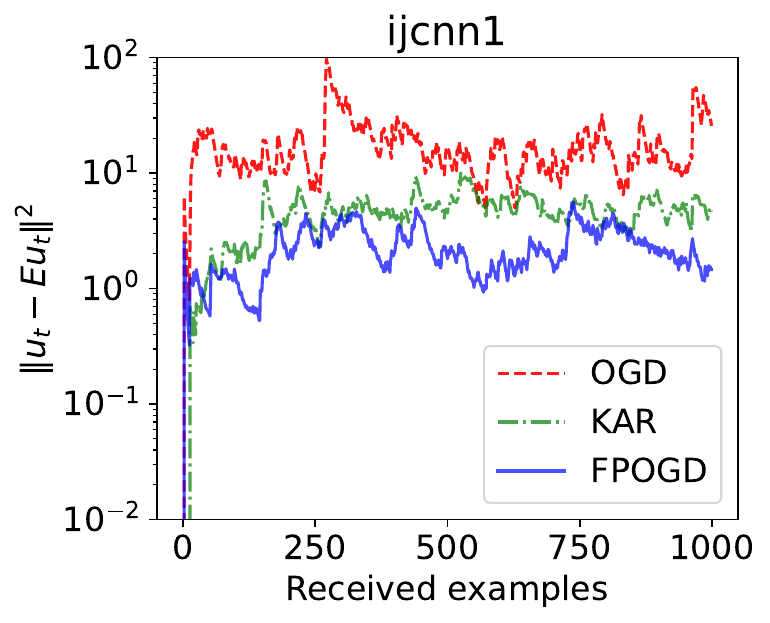}
          \includegraphics[width=0.24\textwidth]{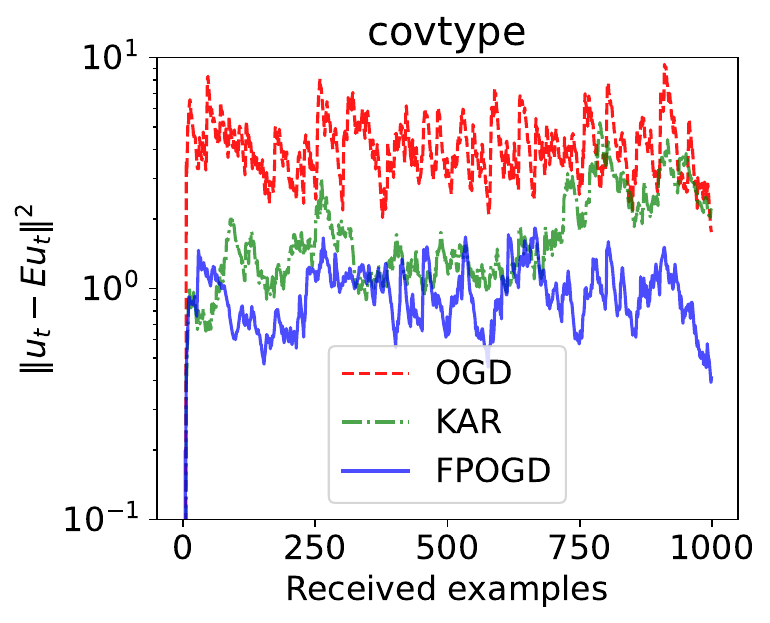}
          
         \includegraphics[width=0.24\textwidth]{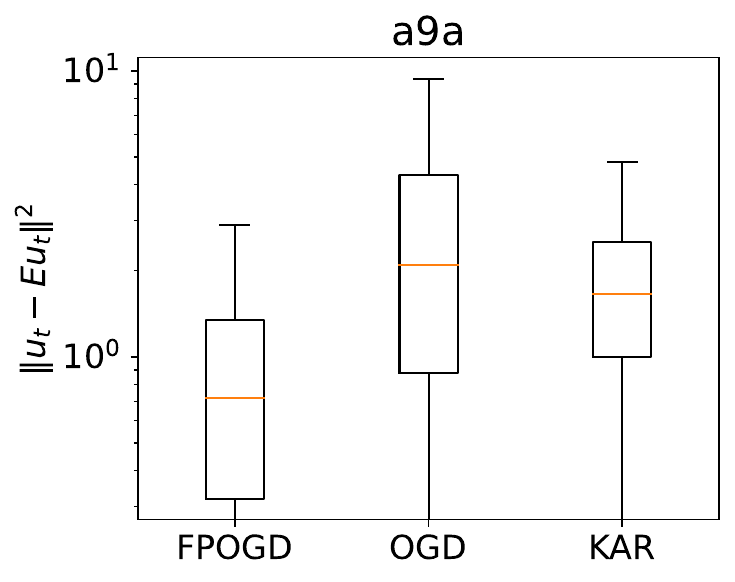}
         \includegraphics[width=0.24\textwidth]{ACML/newplts/usps_varBox.pdf}
         \includegraphics[width=0.24\textwidth]{ACML/newplts/ijcnn1_varBox.pdf}
          \includegraphics[width=0.24\textwidth]{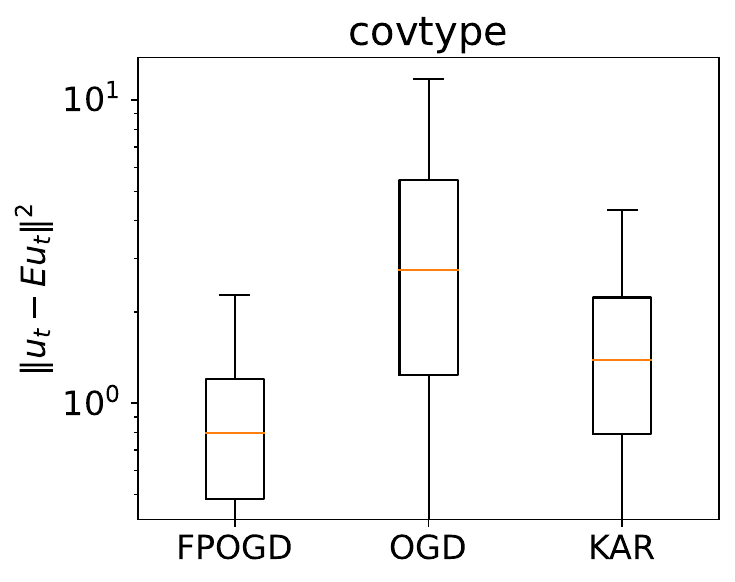}
             \caption{Analyzing Gradient Variance with 4-Size Buffer Across Four Datasets: This study investigates gradient variance using a 4-size buffer across four datasets. The top row shows running variances, while the bottom row presents logarithmic box plots. "Received examples" are online inputs at each time step, with $u_t$ as the online stochastic gradient, and $Eu_t$ as the expected gradient. The red line in the box plots represents the mean variance of stochastic gradients across all algorithms.}
             \centering
             \label{fig:var1}
             \vspace{-5mm}
\end{figure*}

\begin{figure*}        [ht]

     \centering
         \includegraphics[width=0.24\textwidth]{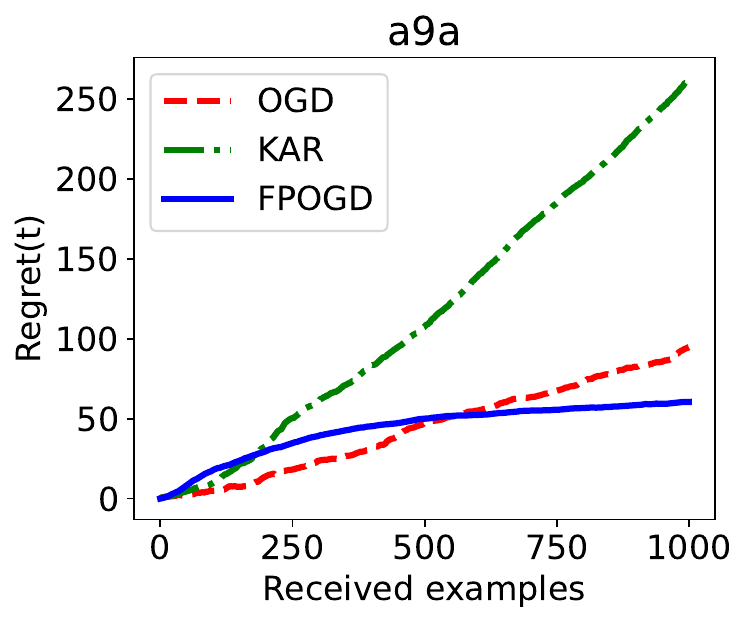}
         \includegraphics[width=0.24\textwidth]{ACML/newplts/usps_regretplot.pdf}
         \includegraphics[width=0.24\textwidth]{ACML/newplts/ijcnn1_regretplot.pdf}
        \includegraphics[width=0.24\textwidth]{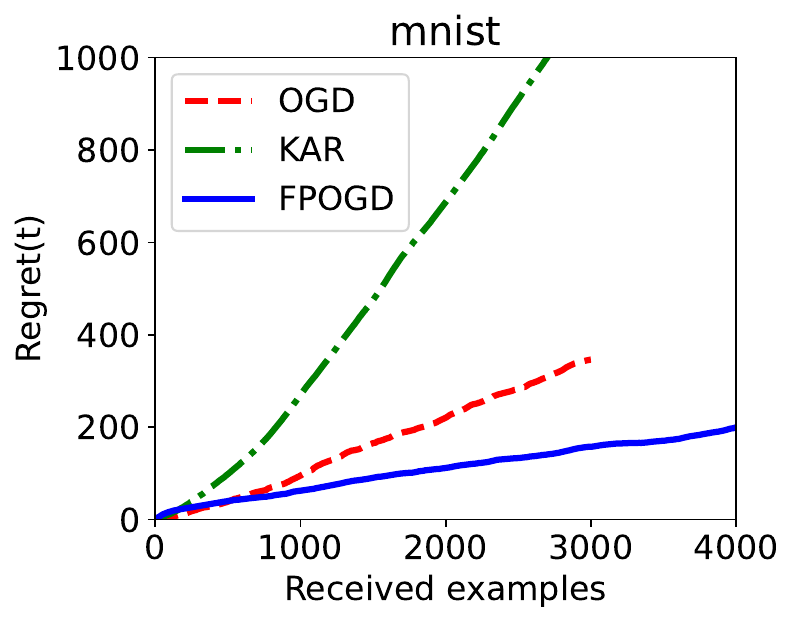}

        \includegraphics[width=0.24\textwidth]{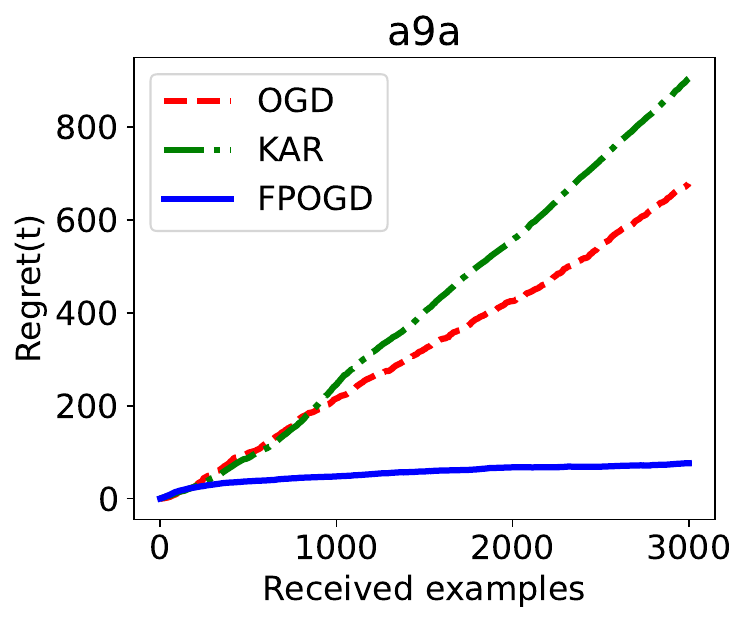}
        \includegraphics[width=0.24\textwidth]{ACML/newplts/usps_regretplot_notiid.pdf}        \includegraphics[width=0.24\textwidth]{ACML/newplts/ijcnn1_regretplot_notiid.pdf}
\includegraphics[width=0.24\textwidth]{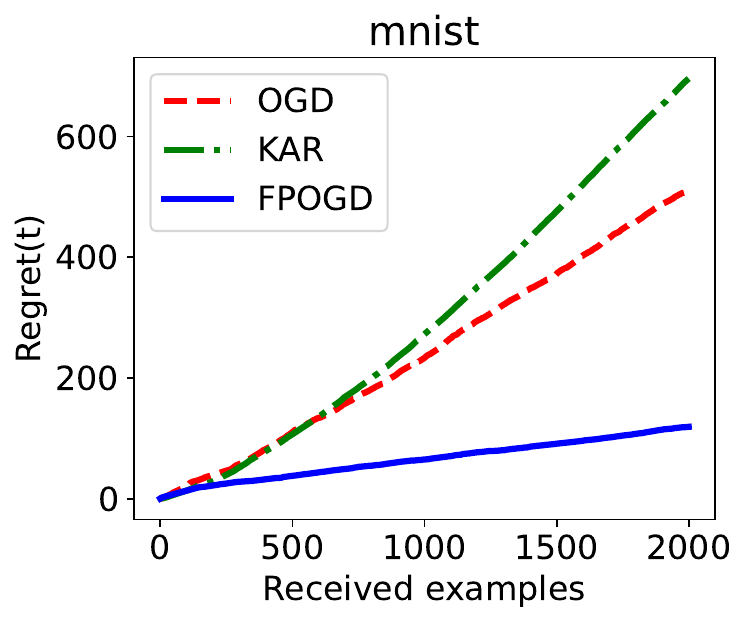}
        \caption{Comparing Regret Across Algorithms and Datasets: This figure illustrates regret, defined as $\sum_t L_t(w_t)-L_t(w^*)$, across diverse algorithms and datasets. The first row depict various algorithms applied to i.i.d. datasets, while the second row features non-i.i.d. datasets generated by sorting examples based on a single feature. Notably, our approach exhibits stronger sublinear regret in the non-i.i.d. setting compared to other algorithms.}
             \centering
             \label{fig:regret1}
             \vspace{-5mm}

\end{figure*}

\begin{figure*}        [ht]

     \centering
         \includegraphics[width=0.31\textwidth]{imgs/usps.pdf}
         \includegraphics[width=0.31\textwidth]{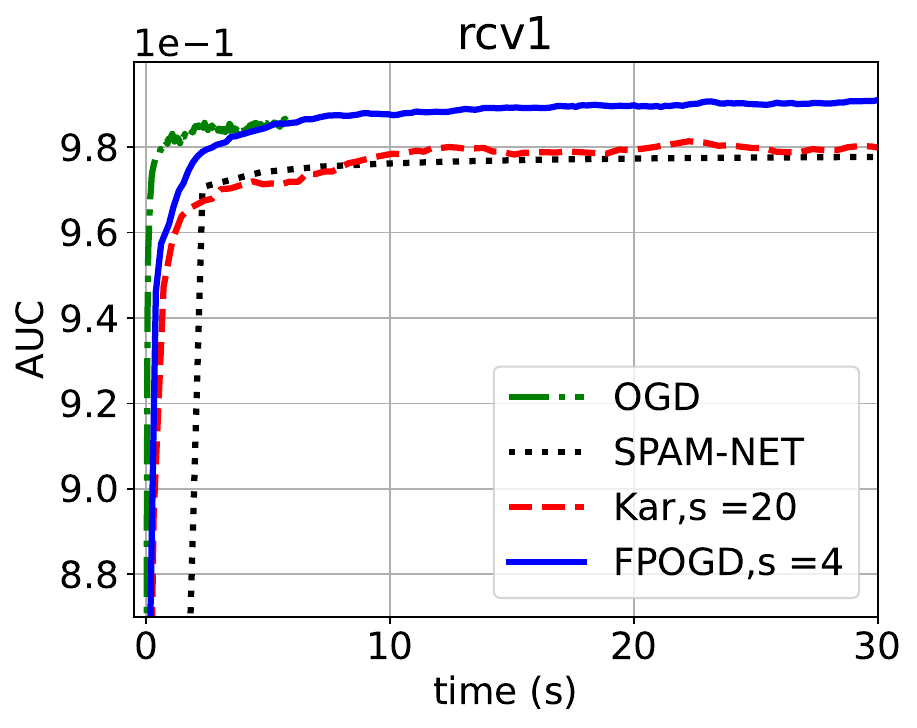}
         \includegraphics[width=0.31\textwidth]{imgs/ijcnn1.pdf}
          \includegraphics[width=0.31\textwidth]{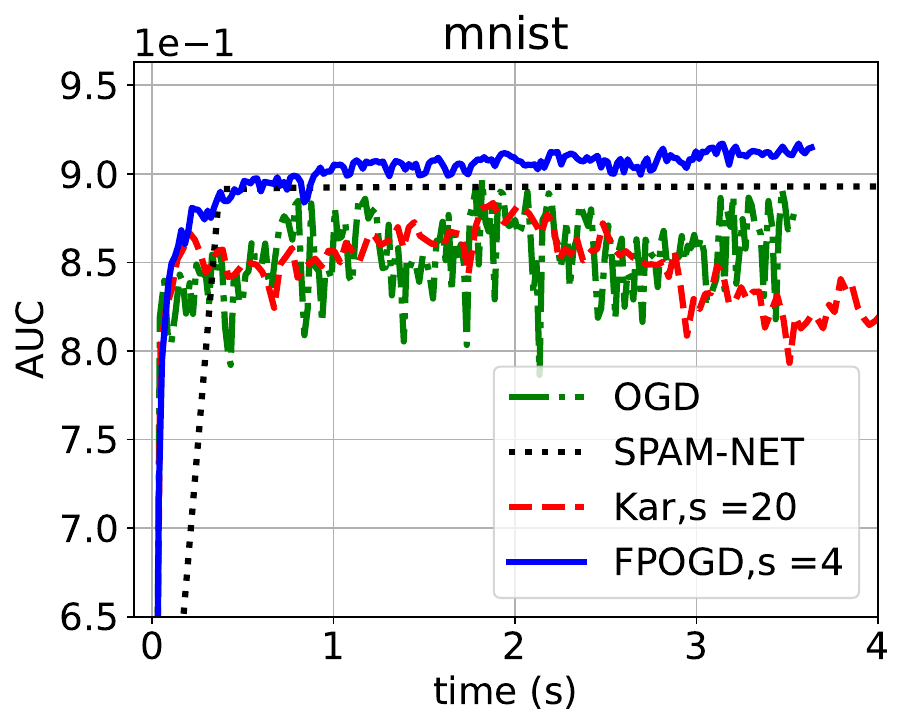}
                    \includegraphics[width=0.31\textwidth]{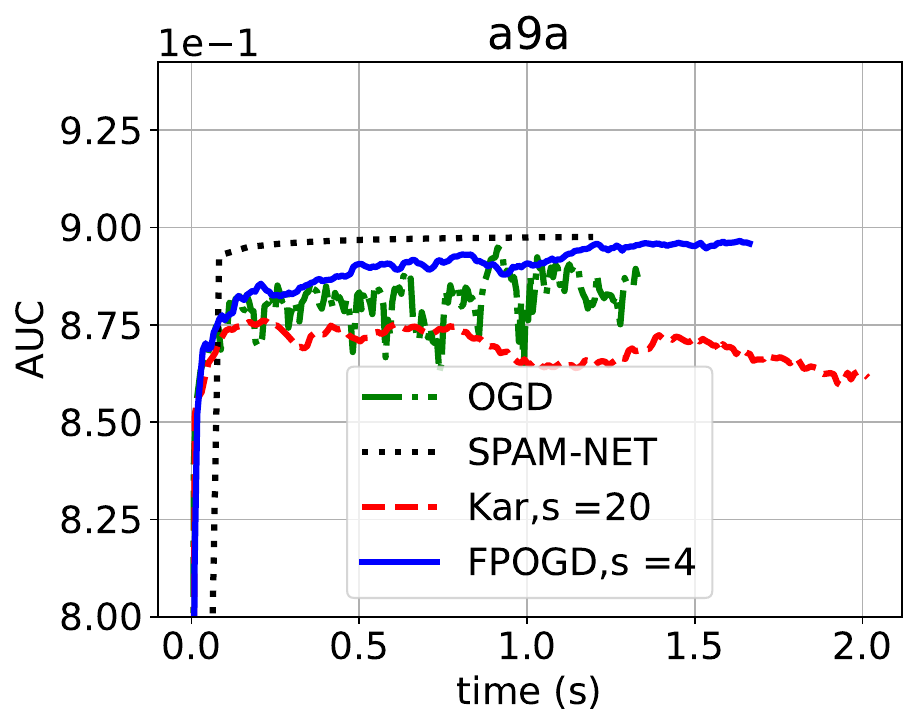}
           \includegraphics[width=0.31\textwidth]{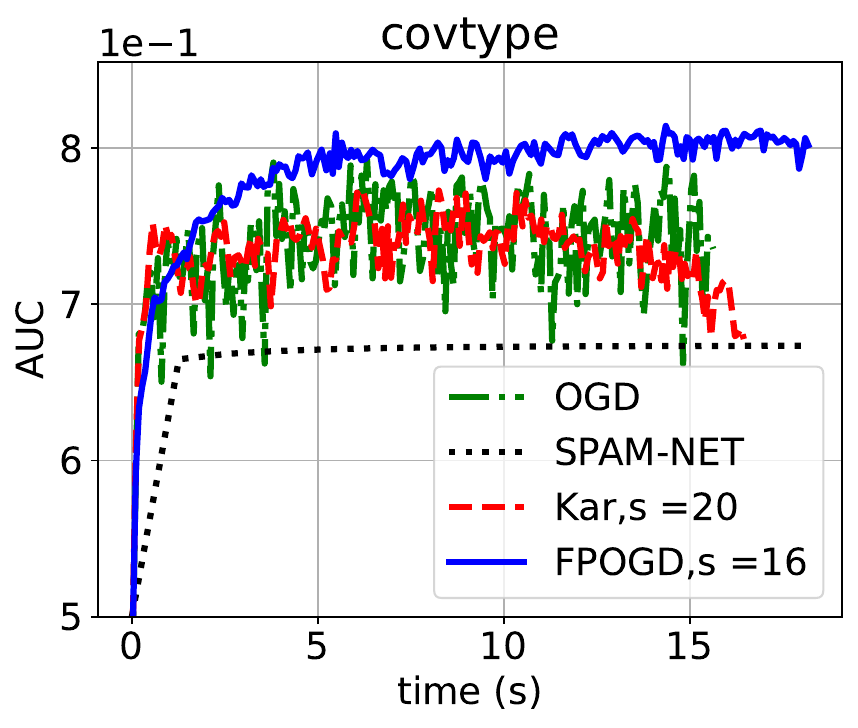}

             \caption{The AUC vs. time comparison of the algorithms in different datasets showing superior performance of the proposed method. }
             \centering
             \label{fig:results1}
             \vspace{-5mm}
\end{figure*}

In this section, we present a series of figures that highlight the superior performance of our method: Figure \ref{fig:results1} demonstrates the efficiency and effectiveness of our approach by showcasing the Area Under the Curve (AUC) in relation to time. Figure \ref{fig:var1} provides insights into gradient variance analysis using a 4-size buffer, emphasizing the stability of our algorithm. Finally, Figures \ref{fig:regret1} compare regret values on i.i.d. and non-i.i.d. datasets, showcasing the adaptability and robustness of our method across different data distribution scenarios.

\end{document}